\pdfoutput=1
\documentclass[twoside,11pt]{article}
\usepackage{amsmath,amssymb,mathtools}
\usepackage{enumitem}
\usepackage{array}
\usepackage{booktabs}
\usepackage{graphicx}
\usepackage{tikz,pgfplots}
\pgfplotsset{compat=1.17}
\usepgfplotslibrary{fillbetween}
\usepackage{jmlr2e}
\renewenvironment{proof}[1][Proof]{\par\noindent{\bf #1\ }}{\hfill\BlackBox\\[2mm]}
\hypersetup{
  pdftitle={The Price of Not Planning: Parallel Sampling Depth Peaks at the
            Reconstruction Threshold},
  pdfauthor={Chuling Wen, Weijie Liang, Jian Lu},
  pdfsubject={Parallel sampling; masked diffusion models; broadcasting on trees},
  pdfkeywords={masked diffusion models, parallel decoding, latent variables,
               reconstruction threshold, serial depth}}

\newcommand{\KL}{\mathrm{KL}}
\newcommand{\TC}{\mathrm{TC}}
\newcommand{\E}{\mathbb{E}}
\newcommand{\R}{\mathbb{R}}
\newcommand{\eps}{\varepsilon}
\DeclareMathOperator{\corr}{corr}
\DeclareMathOperator{\Var}{Var}
\DeclareMathOperator{\Cov}{Cov}
\newcommand{\absdraft}{Masked diffusion models sample the tokens of one
round independently. Planned decoders first generate a coarse plan, then
fill in the tokens in parallel. We ask when a plan saves rounds, how many
it saves, and how fine it must be. We study this on broadcast trees: the
leaves are the tokens, the internal nodes are the plan, and the edge
correlation $\theta$ sets how strongly tokens depend on the plan. Errors
are measured by the forward Kullback--Leibler divergence $\varepsilon$,
and the lower bounds hold for every deterministic value-adaptive rule,
also with learned conditionals. We show that a decoder without the plan
needs of order $\min\{n,(n/\varepsilon)^{\alpha}\}$ rounds below the
reconstruction threshold, where $\alpha=\log b/(4\log(1/\theta))$ rises to
$1/2$ at the threshold, and that the exponent stays below $1/2$ above it
(binary channel: $0.02\le\theta\le0.98$). With the plan, $\log_bn+1$
rounds suffice. In the binary model below the threshold, a plan with units
of $m$ tokens needs of order
$\log_b(n/m)+\min\{m,(n/\varepsilon)^{\alpha}\}$ rounds, so it gives
logarithmic depth exactly when $m$ is logarithmic. A plan therefore helps
most near the threshold, and only if its units are small. \EXPABS}
\newcommand{\EXPABS}{Exact divergence computations in the Gaussian model
with up to $2^{20}$ leaves confirm these exponents below the threshold.}

\newcommand{\EXPABOVE}{Table~\ref{tab:exp-above} compares the multistage
schedules with the stride schedule for $\theta>2^{-1/2}$ on trees with up
to $n=8192$ leaves, with exponents fitted over $h=8,\dots,13$. The
exponent of the stride schedule increases with $\theta$, exceeds $1/2$ for
$\theta\ge0.75$, and reaches $0.80$ at $\theta=0.9$. The multistage
schedules keep the exponent between $0.39$ and $0.48$ and reduce the
depth at $n=8192$ by factors from $1.4$ to $18$.

The peak of Corollary~\ref{cor:peak} is not visible at these sizes. For
$\theta$ between $0.72$ and $0.9$ the fitted exponents of the multistage
schedules show no trend, and for $\theta\ge0.8$ they exceed
$a_{\mathrm{bin}}$. They stay below the local slope of the bound of
Theorem~\ref{thm:binary-above}, which includes its logarithmic factors
(Appendix~\ref{app:exp}). In the Gaussian model the divergence can be
computed without sampling for larger trees, and the decrease is visible
there.}
\newcommand{\EXPABOVEROWS}{0.72 & 0.499 & 0.45 (0.03) & 0.44 (0.05) & 39.6 & 28.9\\ 0.75 & 0.489 & 0.52 (0.02) & 0.39 (0.02) & 70.7 & 34.9\\ 0.80 & 0.447 & 0.62 (0.03) & 0.48 (0.03) & 184 & 60.6\\ 0.85 & 0.379 & 0.70 (0.01) & 0.48 (0.04) & 393 & 81.4\\ 0.90 & 0.307 & 0.80 (0.05) & 0.45 (0.05) & 1180 & 65.1\\}
\newcommand{\EXPRULESROWS}{0.40 & 2.8 & 3.2 & 286 & $>2048$\\ 0.60 & 9.0 & 8.8 & 1124 & $>2048$\\ 0.7071 & 28.6 & 23.8 & 1540 & $>2048$\\ 0.80 & 99.5 & 39.9 & $>2048$ & $>2048$\\ 0.90 & 731 & 55.7 & $>2048$ & $>2048$\\}
\newcommand{\EXPPLAN}{At $\theta=0.6$ every plan needs more rounds than the multistage
schedule without a plan, $11.7$ to $14.6$ against $8.8$, since the leaf
depth is already below $h+1$. At $\theta=2^{-1/2}$ and $\theta=0.8$, plans with units of at most
$16$ leaves need $12$ to $14$ rounds, close to $h+1=13$, against $24$ and
$40$ rounds for the multistage schedule without a plan. Plans with units of $256$ or more
leaves are worse than the multistage schedule at $\theta=0.8$, because the leaves
inside each unit are generated by the stride schedule. A plan therefore
saves rounds when the leaf depth exceeds $h+1$, which at $n=4096$ happens
near and above the threshold.}
\newcommand{\EXPPLANROWS}{0.60 & 8.8 & 11.7 & 13.7 & 14.6 & 13.3 & 12.2 & 13.0\\ 0.7071 & 23.8 & 29.9 & 27.5 & 20.7 & 13.5 & 12.3 & 13.0\\ 0.80 & 39.9 & 72.4 & 44.7 & 23.9 & 13.7 & 12.3 & 13.0\\}

\ShortHeadings{The Price of Not Planning}{Wen, Liang and Lu}
\firstpageno{1}
\makeatletter
\def\@starteditor{}
\def\@endeditor{}
\makeatother
\editor{}

\begin{document}

\title{The Price of Not Planning: Parallel Sampling Depth\\
Peaks at the Reconstruction Threshold}

\author{\name Chuling Wen \email chulwen@163.com \\
        \addr Shenzhen Key Laboratory of Advanced Machine Learning and
        Applications\\
        School of Mathematical Sciences, Shenzhen University\\
        Shenzhen 518060, China
        \AND
        \name Weijie Liang \email liangweijie2022@email.szu.edu.cn \\
        \addr Shenzhen Key Laboratory of Advanced Machine Learning and
        Applications\\
        School of Mathematical Sciences, Shenzhen University\\
        Shenzhen 518060, China
        \AND
        \name Jian Lu\thanks{Corresponding author.} \email jianlu@szu.edu.cn \\
        \addr Shenzhen Key Laboratory of Advanced Machine Learning and
        Applications\\
        School of Mathematical Sciences, Shenzhen University\\
        Shenzhen 518060, China\\
        National Center for Applied Mathematics Shenzhen (NCAMS)\\
        Shenzhen 518055, China}

\maketitle

\begin{abstract}
\absdraft
\end{abstract}


\section{Introduction}\label{sec:intro}

Masked diffusion models generate a sequence in rounds
\citep{austin2021,radd2024,llada2025}. Each round is one model call that
fills several masked positions, and the positions of a round are sampled
independently from the model's conditionals given the positions filled so
far. Even when every conditional is exact, the joint draw then loses the
dependence among them, and generation quality deteriorates as more
positions are filled together \citep{parallelbench2026}. A common remedy is
to generate a plan first. Planned diffusion writes a short plan that splits
the response into chunks meant to be independent and then denoises the
chunks in parallel \citep{israel2026}. Other models generate latent
variables before or alongside the tokens \citep{hu2026,draxler2026}. When
does a plan save rounds, how many does it save, and how fine must it be?

For data whose tokens are the only variables, the minimum number of
rounds, the serial depth, is known for several families \citep{wen2026}.
In a Markov chain the generated tokens separate the remaining ones, so a
decoder can place widely spaced anchors and fill the gaps between them in
parallel \citep{bansal2026}. Conditioning on separators plays the same
role in Gaussian lattice fields \citep{chenyeh2026}. Hierarchical data
behave differently. Two tokens remain dependent through every common
ancestor that has not been generated, and generated tokens never separate
them. Only the latent ancestors do, and only a decoder that may generate
them can use them. Existing results therefore do not say what a plan buys.

We study the simplest hierarchical source, broadcasting on a tree. Here
the tokens depend on one another only through their latent ancestors, so a
plan is the only way to remove their dependence. The conditional laws and
the reconstruction threshold are known exactly, so the rounds a plan saves
can be determined rather than estimated. Fix
integers $b\ge2$ and $h\ge1$, and let $T_{b,h}$ be the complete $b$-ary
tree of height $h$ with $n=b^h$ leaves. In the binary model,
\begin{equation}\label{eq:binary-model}
 Z_o\sim\mathrm{Unif}\{\pm1\},\qquad
 Z_v=Z_{\mathrm{pa}(v)}\,\eta_v,\qquad
 \Pr(\eta_v=1)=\tfrac{1+\theta}{2},
\end{equation}
where $o$ is the root, $\mathrm{pa}(v)$ is the parent of $v$, the flips
$\eta_v\in\{\pm1\}$ are independent, and $\theta\in(0,1)$ is the edge
correlation. In the Gaussian model,
\begin{equation}\label{eq:gauss-model}
 Z_o\sim N(0,1),\qquad Z_v=\rho Z_{\mathrm{pa}(v)}+\sqrt{1-\rho^2}\,\xi_v,
\end{equation}
where the $\xi_v$ are independent standard Gaussian variables and
$\rho\in(0,1)$. Table~\ref{tab:dictionary} lists how the model corresponds
to plan-then-fill decoding. Tree-structured generative models are a common
setting for theoretical questions about hierarchical data. They have been
used to study the role of depth in learning such data
\citep{mossel2016,cagnetta2024}. Image diffusion models show the phase
transition that these models predict \citep{sclocchi2025phase}, and masked
diffusion language models and image diffusion models show the predicted
correlated changes \citep{sclocchi2025probing}. A single parameter, the edge correlation, sets how much the leaves reveal
about their ancestors. This lets us vary the role of the latent variables
continuously.

\begin{table}[t]
\centering
\small
\begin{tabular}{@{}ll@{}}
\toprule
Broadcast model & Plan-then-fill decoding\\
\midrule
leaves $X_1,\dots,X_n$ & output tokens\\
internal nodes & plan: outline, latent code or coarse scales\\
leaves below a finest plan node, $m$ & size of a plan unit\\
edge correlation $\theta$ or $\rho$ & how strongly the tokens depend on the plan\\
rule on the leaves & masked diffusion sampler without a plan\\
rule on all nodes & decoder that generates the plan first\\
forward KL at most $\eps$ & error budget for the output distribution\\
\bottomrule
\end{tabular}
\caption{Correspondence between the broadcast model and plan-then-fill
decoding.}
\label{tab:dictionary}
\end{table}

A deterministic rule chooses the next batch of leaves from the values
generated so far and samples the batch independently from its exact
conditional marginals. Its leaf depth $D_\eps$ is the fewest rounds, on
every history, for which some rule keeps the forward Kullback--Leibler
(KL) divergence from the leaf law within $\eps$
(Definition~\ref{def:depth}). By the cost identity of
\citet{benhamu2025}, this divergence is the expected conditional total
correlation of the batches, summed over the rounds
(Proposition~\ref{prop:cost}). Two leaves whose nearest common ancestor has
height $d$ share mutual information of order $\theta^{4d}$ when nothing
else is known, which suggests a depth of order $(n/\eps)^{\alpha}$ with
\begin{equation}\label{eq:alpha}
 \alpha=\frac{\log b}{4\log(1/\theta)} .
\end{equation}
The difficulty is that generated leaves carry information about their
ancestors. When $b\theta^2>1$, the Kesten--Stigum condition, many leaves
together determine an ancestor with nonvanishing accuracy, and the count
above no longer holds.

We determine the leaf depth of broadcast trees below the reconstruction
threshold for all deterministic value-adaptive rules, bound it above the
threshold, and show that a decoder that generates the latent tree needs
only $h+1$ rounds. Our contributions are as follows.
\begin{itemize}
\item \emph{The price of not planning.} Below the reconstruction threshold
$b\theta^2<1$, the leaf depth is of order $\min\{n,(n/\eps)^{\alpha}\}$
in both models (Theorems~\ref{thm:binary-below} and~\ref{thm:gauss}), and
$\alpha$ increases to $1/2$ at the threshold. Above it, schedules that
generate sparse leaves before dense ones keep the exponent below $1/2$, in
the Gaussian model and in the binary model for $0.02\le\theta\le0.98$
(Theorems~\ref{thm:gauss} and~\ref{thm:binary-above}). The binary proof
reduces to a one-step inequality along tree paths, which we verify by
interval arithmetic. The Gaussian exponent therefore peaks at the
threshold. The binary exponent rises to $1/2$ below the threshold and stays
below $1/2$ above it in the verified range (Corollary~\ref{cor:peak}). A
plan thus saves the most rounds near the threshold: weakly correlated data
need few rounds without a plan, and strongly correlated leaves carry an
implicit plan.
\item \emph{What a plan buys and how fine it must be.} A decoder that
generates the internal nodes samples exactly in $h+1$ rounds, and in the
binary model logarithmically many rounds are necessary for the joint law
(Theorem~\ref{thm:plan}). In the binary model below the threshold, a plan
whose finest units cover $m$ leaves needs of order
$\log_b(n/m)+\min\{m,(n/\eps)^{\alpha}\}$ rounds
(Theorem~\ref{prop:granularity}). A plan therefore reaches logarithmic
depth exactly when its units have logarithmic size.
\item \emph{How to order the tokens without a plan.} The schedules that
attain the upper bounds put leaves that are far apart in the tree into the
same round and generate sparse leaves before dense ones. Computations
with exact conditionals on binary trees with up to $8192$ leaves and
Gaussian trees with up to $2^{20}$ leaves match the predicted exponents
below the threshold and show that staged schedules reduce the depth above it
(Section~\ref{sec:experiments}). Uniformly random batches and
confidence-based selection follow neither rule and need a number of
rounds that grows about linearly in $n$.
\end{itemize}
The lower bounds also hold for learned conditionals
(Proposition~\ref{prop:learned}). Table~\ref{tab:depths} summarizes the
bounds and the schedules that attain them.

\begin{table}[t]
\centering
\footnotesize
\setlength{\tabcolsep}{3pt}
\begin{tabular}{>{\raggedright\arraybackslash}p{0.2\textwidth}>{\raggedright\arraybackslash}p{0.3\textwidth}>{\raggedright\arraybackslash}p{0.31\textwidth}>{\raggedright\arraybackslash}p{0.13\textwidth}}
\hline
regime & proved depth bounds & schedule attaining the upper bound & source\\
\hline
leaves only, $b\theta^2<1$ (both models) & $D_\eps\asymp\min\{n,(n/\eps)^{\alpha}\}$ & one leaf of every subtree of height $d$ per round, $b^d$ rounds & Theorems~\ref{thm:binary-below}, \ref{thm:gauss}\\
leaves only, Gaussian, $b\rho^2=1$ & $\min\{n,\sqrt{n/\eps}/h^2\}\lesssim D_\eps\lesssim\min\{n,\sqrt{n/\eps}\}$ & as above & Theorem~\ref{thm:gauss}\\
leaves only, Gaussian, $b\rho^2>1$ & $\min\{n,(n/\eps)^{\alpha_{\mathrm{lo}}}\}\lesssim D_\eps\lesssim\min\{n,h(nh^2/\eps)^{\alpha_{\mathrm{up}}}\}$, both exponents $<1/2$, equal when $b\rho^4\ge1$ & sparse leaves first, then denser ones, in stages & Theorem~\ref{thm:gauss}\\
leaves only, binary, $b\theta^2>1$, $\theta\in[0.02,0.98]$ & $D_\eps\lesssim h(nh^2/\eps)^{\gamma}$ for every $\gamma>a_{\mathrm{bin}}$, $a_{\mathrm{bin}}<1/2$ & as above & Theorem~\ref{thm:binary-above}\\
all nodes (plan generated) & $D^{\mathrm{all}}_0\le h+1$; $D^{\mathrm{all}}_\eps\gtrsim\log n$ (binary) & one level of the tree per round & Theorem~\ref{thm:plan}\\
plan down to subtrees of $m$ leaves, binary, $b\theta^2<1$ & $D^{(s)}_\eps\asymp(h-s+1)+\min\{m,(n/\eps)^{\alpha}\}$ & plan level by level, then the first schedule inside each subtree & Theorem~\ref{prop:granularity}\\
\hline
\end{tabular}
\caption{Depths and attaining schedules. Lower bounds hold for all
deterministic value-adaptive rules with exact conditionals and remain
valid for learned conditionals; all schedules are fixed in advance. Here
$\alpha=\log b/(4\log(1/\theta))$, $\alpha_{\mathrm{lo}}$ and
$\alpha_{\mathrm{up}}$ are given in Theorem~\ref{thm:gauss}, and
$a_{\mathrm{bin}}$ in Theorem~\ref{thm:binary-above}. Constants depend only
on $b$ and the edge correlation, and in the binary row above the threshold
also on $\gamma$.}
\label{tab:depths}
\end{table}

Section~\ref{sec:model} defines the broadcast models, the rules and the
depth. Section~\ref{sec:cost} states the cost identity, extends it to
learned conditionals, and shows how it yields depth bounds.
Section~\ref{sec:leaf} derives the leaf depth below and above the
reconstruction threshold, and Section~\ref{sec:plan} treats decoders that
generate the plan. Section~\ref{sec:experiments} reports experiments, and
Section~\ref{sec:conclusion} concludes. Proofs are in the appendices.

\subsection{Related Work}\label{sec:related}

Prior work concerns decoding with plans and latent variables, error
decompositions and depth bounds for parallel decoding, and reconstruction
on trees; we discuss them in this order.
Planned diffusion generates a plan autoregressively and then denoises its
chunks in parallel \citep{israel2026}. Representation-based masked
diffusion conditions the tokens on a latent representation and relates
the conditional total correlation of a batch with and without the latent
variable \citep[Propositions~3.1 and~3.2]{hu2026}. Variational decoding adds token-level auxiliary
latent variables so that one forward pass generates many tokens
\citep{draxler2026}. These works propose and evaluate mechanisms. We
specify which latent coordinates the decoder may generate and quantify the
rounds saved.
\citet[Theorems~3.1 and~3.4]{gaitonde2026} study sequential autoregressive
generation of the leaves of the binary broadcast process. Above the
reconstruction threshold, the sum of the leaves generated by their
bounded-context autoregressive process has a polynomially smaller variance
than under the true law, unless the context covers a constant fraction of
the leaves. A working memory of $O(\log n)$ bits that stores the ancestors
suffices for exact sampling. Their resource is the memory of a sequential
sampler. Ours is the number of parallel rounds, and we bound it for every
deterministic value-adaptive rule.

\citet[Sec.~3.1, Eqs.~(4)--(5)]{halton2025} express the cost of a step of
a masked generative image transformer as the divergence from the joint
conditional to the product of its marginals, sum it over steps, and choose
a Halton order by spatial separation.
\citet{benhamu2025} decompose the forward KL error of an adaptive rule into
model error and the conditional total correlation of its batches, with
equality for deterministic rules; Proposition~\ref{prop:cost} is the exact
case. \citet{wen2026} define serial depth as the fewest rounds over all
deterministic value-adaptive rules and determine it for Markov chains and
other families without latent variables. \citet{bansal2026} study
parallel samplers through random walks on graphs; their bisection
samplers take logarithmically many steps and are exact for random walks
when the marginals are exact. \citet{li2026random} show that at every
fixed number of rounds of at least two, randomized schedules can be far
more accurate than deterministic value-adaptive rules, so lower bounds for
deterministic rules do not extend to randomized samplers.
\citet[Theorem~3.2]{fu2025bits} show that a decoder whose confidence
threshold tightens with the batch size, as in their Assumption~3.1, needs a number of rounds that grows linearly with
the total information of the sample and inversely with the information
budget per round. \citet{kou2026} prove polynomial lower bounds on the
number of adaptive rounds of diffusion sampling in $\R^d$ with approximate
scores for certain target families, for randomized algorithms that make polynomially many score
queries per round. \citet{chen2025} optimize batch
sizes under uniformly random selection of positions. \citet{chenyeh2026}
study product-round sampling on observed Gaussian lattice fields, where
conditioning on separators removes the remaining dependence.
\citet{wucai2026} give learning and sampling guarantees for masked
diffusion models on mixtures of hidden Markov trees with random batch
schedules. \citet{sheriff2026} gives end-to-end guarantees for learned
denoisers on stationary sources under conditional mixing, where revealing
distant tokens moves the posterior of a masked token exponentially little
in the distance. With correlation length $\tau_c$, sequence length $N$
and vocabulary $\mathcal V$, a dilated schedule then needs
$O(\tau_c\log(N|\mathcal V|/\eps))$ steps. The leaves of a broadcast tree
are not stationary. With the leaves numbered from $0$, the leaves $0$ and
$k$ have their nearest common ancestor at height $\lceil\log_b(k+1)\rceil$
and correlation $\theta^{2\lceil\log_b(k+1)\rceil}$, which decays only
polynomially in $k$. In
our setting the generated leaves are not separators of the hidden tree,
and we determine how the depth exponent depends on the edge correlation.
Three classes of samplers are outside the one-shot rules studied here:
samplers that verify and correct speculative configurations
\citep{anari2024,anari2025}, rejection samplers that correct independent
draws \citep{zhang2026sirs}, and parallel Glauber-type dynamics that
resample coordinates \citep{liuyin2025}.

The reconstruction problem asks whether the leaves of a broadcast process
carry nonvanishing information about the root. For the binary channel,
the probability of correct reconstruction tends to $1/2$ when
$b\theta^2<1$ and stays bounded away from $1/2$ when $b\theta^2>1$
\citep{ekps2000}; $b\theta^2=1$ is known as the Kesten--Stigum bound
\citep{kestenstigum1966}. The Gaussian channel has the analogous threshold
$b\rho^2=1$: the fraction of the variance of the root explained by the
leaves has a positive limit exactly when $b\rho^2>1$ (proof of Lemma~\ref{lem:gauss-floor}). Our leaf-only lower bounds use the
corresponding decay of information below the threshold, which we prove in
the form needed for value-adaptive conditioning (Appendix~\ref{app:binary-below}).
Above the threshold the depth depends on how fast the information about an
ancestor saturates as leaves are generated, a quantity that does not appear
in the reconstruction problem itself.

\section{Broadcast Trees and Reveal Rules}\label{sec:model}

This section sets up the objects that the rest of the paper bounds. A
decoder is modeled as a rule that, after each round, sees the values
generated so far and chooses the next batch, which is then sampled
independently from exact conditional distributions. The depth is the
smallest number of rounds such a rule needs to keep the forward KL error
within a budget. Section~\ref{sec:notation} fixes the notation and
defines the conditional total correlation,
Section~\ref{sec:trees} fixes the parameters of the two
broadcast models, and Section~\ref{sec:rules} defines rules and depth for
leaf-only decoders and for decoders that may generate the plan.

\subsection{Notation and Preliminaries}\label{sec:notation}

The complete $b$-ary tree $T_{b,h}$ has root $o$ and $n=b^h$ leaves, and
$\mathrm{pa}(v)$ is the parent of a node $v$. The height of a node is its
distance to the leaves, so leaves have height $0$ and the root has height
$h$. Constants $c,C$ depend only on $b$ and the edge correlation, not on
$h$, $n$ or $\eps$.

For a random vector $X=(X_i)_{i\in V}$ on a finite index set $V$ and
disjoint $S,R\subseteq V$, the conditional total correlation
\[
 \TC(X_S\mid x_R)=\sum_{i\in S}H(X_i\mid x_R)-H(X_S\mid x_R)
\]
is nonnegative and vanishes exactly when the coordinates in $S$ are
conditionally independent given $x_R$; for the Gaussian model the entropies
are differential entropies. It measures what a batch loses when it is
sampled independently. It is superadditive over every partition of $S$,
and for every ordering $s_1,s_2,\dots$ of $S$ it equals
$\sum_{i\ge2}I(X_{s_i};X_{s_1},\dots,X_{s_{i-1}}\mid x_R)$
(Lemma~\ref{lem:tc} in Appendix~\ref{app:cost}).

\subsection{Broadcast Processes}\label{sec:trees}

The binary model~\eqref{eq:binary-model} and the Gaussian
model~\eqref{eq:gauss-model} live on $T_{b,h}$. Two leaves whose nearest
common ancestor has height $d$ are at distance $2d$, and their correlation
is $\theta^{2d}$, respectively $\rho^{2d}$. We write
\[
 x=\frac{\log(1/\theta)}{\log b}\quad\text{(binary)},\qquad
 x=\frac{\log(1/\rho)}{\log b}\quad\text{(Gaussian)},
\]
so that the edge correlation is $b^{-x}$. The reconstruction
(Kesten--Stigum) threshold $b\theta^2=1$, respectively $b\rho^2=1$, is
$x=1/2$, and the exponent~\eqref{eq:alpha} is $\alpha=1/(4x)$. Below the
threshold $x>1/2$ and $\alpha<1/2$; above it $x<1/2$.

\subsection{Reveal Rules and Depth}\label{sec:rules}

Let $P$ be the law of $X=(X_i)_{i\in V}$. In the tree models, $V$ is
either the set of leaves or the set of all nodes.

\begin{definition}[Deterministic one-shot rule]\label{def:rule}
A \emph{reveal state} is $(R,x_R)$ with $R\subseteq V$. A
\emph{deterministic value-adaptive rule} $A$ maps each state with $R\ne V$
to a nonempty set $A(R,x_R)\subseteq V\setminus R$. Starting from
$R_1=\emptyset$, round $t$ chooses $S_t=A(R_t,x_{R_t})$, draws
\begin{equation}\label{eq:product-rule}
 Q(dx_{S_t}\mid x_{R_t})=\prod_{i\in S_t}P(dx_i\mid x_{R_t}),
\end{equation}
and sets $R_{t+1}=R_t\cup S_t$. We write $Q_A$ for the law of the output.
\end{definition}

Each coordinate is generated exactly once and is never revised. A
\emph{fixed schedule} chooses each batch from the round index alone.
Exact conditional marginals isolate the error of within-round independence
from prediction error; Proposition~\ref{prop:learned} adds the latter.

\begin{definition}[Depth]\label{def:depth}
Let $\mathcal A_D(V)$ be the set of deterministic value-adaptive rules that
use at most $D$ rounds on every history. For $\eps\ge0$,
\[
 D_\eps(P;V)=\min\bigl\{D\ge1:\ \text{some }A\in\mathcal A_D(V)\text{ has }\KL(P\,\|\,Q_A)\le\eps\bigr\}.
\]
For the tree models, the \emph{leaf depth} $D_\eps$ takes $V$ to be the
leaves and $P$ the leaf law, and the \emph{full depth}
$D^{\mathrm{all}}_\eps$ takes $V$ to be all nodes and $P$ the joint law.
\end{definition}

Generating one coordinate per round is exact, so $D_0\le n$, and both
depths are nonincreasing in $\eps$. A rule on all nodes yields a leaf
sample by discarding the internal nodes, and the leaf divergence is at most
the joint divergence. The full depth therefore counts the rounds of a
decoder that writes the plan before or during the tokens. It also
requires the plan itself to be drawn correctly.

\begin{remark}[Hard cap and randomized rules]\label{rem:randomized}
The depth caps the number of rounds on every history rather than on
average; all our upper bounds use fixed schedules, whose round count is
constant. With an independent seed $\omega$, a randomized rule has output
law $\E_\omega Q_{A_\omega}$, and convexity gives
$\KL(P\,\|\,\E_\omega Q_{A_\omega})\le\E_\omega\KL(P\,\|\,Q_{A_\omega})$.
A mixture can therefore be better than each of its components, and on
some targets it is far more accurate than every deterministic rule with
the same number of rounds \citep{li2026random}. As in
\citet{wen2026}, our lower bounds concern deterministic rules. Uniformly
random batches, used by most masked diffusion samplers, form a randomized
rule. Section~\ref{sec:experiments} reports their divergence averaged over
seeds, which upper-bounds the divergence of the mixture.
\citet{wu2026mixing} shows that the two criteria can have different round
complexities: for the uniform distribution on balanced binary strings of
length $n$, the least expected number of rounds at a fixed tolerance is of
order $(\log n)^2$ for the schedule-averaged divergence and about
$\log_2\ln n$ for the divergence of the mixture, and the latter is attained
with the same number of rounds on every run.
\end{remark}

\section{The Cost of a Round}\label{sec:cost}

The depth is a minimum over all rules, so bounding it requires a way to
evaluate the error of any rule. This section supplies that tool and shows
how it yields depth bounds on trees. Section~\ref{sec:identity} states the
cost identity and its extension to learned conditionals.
Section~\ref{sec:counting} explains how pairs of leaves that share an
undetermined ancestor turn the identity into lower bounds, and why the
reconstruction threshold changes the count.

\subsection{Exact and Learned Conditionals}\label{sec:identity}

The error of a rule with exact conditionals is the conditional total
correlation of Section~\ref{sec:notation}, accumulated over its rounds.

\begin{proposition}[Cost identity; \citealp{benhamu2025,wen2026}]\label{prop:cost}
For every deterministic one-shot rule $A$,
\[
 \KL(P\,\|\,Q_A)=\E_P\Bigl[\sum_{t}\TC\bigl(X_{S_t}\,\big|\,X_{R_t}\bigr)\Bigr].
\]
\end{proposition}

A trained model replaces the exact conditionals $P(dx_i\mid x_R)$ in
\eqref{eq:product-rule} by predictions $\widehat P(dx_i\mid x_R)$. The rule
may also use the predictions, for example to rank positions by confidence,
as long as it remains a deterministic function of the generated values.
Write $\widehat Q_A$ for the resulting output law.

\begin{proposition}[Learned conditionals]\label{prop:learned}
For every deterministic rule $A$ that uses learned conditionals,
\[
 \KL(P\,\|\,\widehat Q_A)=\E_P\Bigl[\sum_t\Bigl(\TC\bigl(X_{S_t}\,\big|\,X_{R_t}\bigr)
 +\sum_{i\in S_t}\KL\bigl(P(\cdot\mid X_{R_t})\,\big\|\,\widehat P_i(\cdot\mid X_{R_t})\bigr)\Bigr)\Bigr].
\]
In particular $\KL(P\,\|\,\widehat Q_A)\ge\KL(P\,\|\,Q_A)$, where $Q_A$ uses
the same batches with exact conditionals. Every depth lower bound in this
paper therefore holds for decoders with learned conditionals.
\end{proposition}

The first statement is the decomposition of \citet{benhamu2025} for
deterministic rules, and the second follows because the model-error terms
are nonnegative and the batches $S_t$ depend only on the generated values
(Appendix~\ref{app:cost}).

\subsection{From Pairs of Leaves to Depth Bounds}\label{sec:counting}

Every term in the cost identity is nonnegative. A rule therefore meets the
budget $\eps$ only if the dependence it loses over all its rounds is at
most $\eps$ on average. A batch loses at least the sum of the conditional
mutual informations $I(X_i;X_j\mid x_R)$ over any disjoint pairs of its
coordinates, because total correlation is superadditive over a partition
of the batch. Few rounds force large batches, and a
large batch of leaves must contain many pairs that share a low common
ancestor.

On the tree, the conditional dependence of two leaves is carried by the
path between them. If nothing near the path is known, two leaves whose
nearest common ancestor has height $d$ have correlation $\theta^{2d}$ and
mutual information of order $\theta^{4d}$. Suppose that each round
generates one leaf of every subtree of height $d$. A round then costs about
$n b^{-d}\theta^{4d}$, there are $b^d$ rounds, and the total error is about
$n\theta^{4d}$. Reaching error $\eps$ requires
$b^d\approx(n/\eps)^{\alpha}$. Two facts make this count rigorous below the
threshold. For the upper bound, the leaves generated in earlier rounds can
only weaken the dependence along the path, by a contraction that holds for
every conditioning (Appendix~\ref{app:binary-below}). For the lower bound, a
value-adaptive rule could try to concentrate its rounds where the generated
leaves have already pinned down the ancestors. Below the threshold this is
rare: the information that any set of leaves carries about an ancestor of
height $k$ decays like $(b\theta^2)^k$, so few ancestors are ever pinned
down, whatever the rule does.

Above the threshold the second fact fails, and the first becomes the
interesting one. Many generated leaves determine their ancestors to
constant accuracy, and the path between two later leaves passes through
these nearly known ancestors. The correlation along such an edge is smaller
than $\theta$, so dependence decays faster than the count assumes. A rule
that generates sparse leaves first lets later, denser batches see their
ancestors without ever generating them. Section~\ref{sec:leaf} quantifies
how fast the ancestors saturate and what this gains.

\section{Leaf-Only Generation}\label{sec:leaf}

This section derives the depth of decoders that may generate only leaves.
Section~\ref{sec:below} treats the regime below the reconstruction
threshold, where the pair count of Section~\ref{sec:counting} is exact.
Section~\ref{sec:above} treats the regime above it, where generated leaves
act as an implicit plan. Section~\ref{sec:peak} combines the two regimes.

\subsection{Below the Reconstruction Threshold}\label{sec:below}

\begin{theorem}[Binary tree below the threshold]\label{thm:binary-below}
Let $\alpha=\log b/(4\log(1/\theta))$. If $b\theta^2<1$, then there are
constants $c,C>0$ such that, for all $h\ge1$ and $0<\eps\le1/2$,
\[
 c\,\min\{n,(n/\eps)^{\alpha}\}\ \le\ D_\eps\ \le\ C\,\min\{n,(n/\eps)^{\alpha}\}.
\]
The lower bound holds for every deterministic value-adaptive rule; the
upper bound is attained by the fixed schedule that generates one leaf of
every subtree of height $d$ per round, with $b^d\asymp\min\{n,(n/\eps)^{\alpha}\}$.
\end{theorem}

The depth is thus a power of $n$ with exponent $\alpha<1/2$. The exponent
depends on the edge correlation only through $x=\log(1/\theta)/\log b$, and
it increases to $1/2$ as the threshold is approached. Weakly correlated
leaves need few rounds because their dependence is weak; leaves near the
threshold need about $\sqrt{n/\eps}$ rounds. The bound holds at every
fixed $\eps$, so the factor $n^{1-\alpha}$ saved by parallel generation
does not depend on how the budget is chosen. The Gaussian model has the
same exponent below its threshold (Theorem~\ref{thm:gauss}).

For the upper bound of Theorem~\ref{thm:binary-below}, take two leaves
$i,j$ of one batch. Conditionally on
any generated leaves, the pair $(X_i,X_j)$ is a binary Markov chain along
the path through the unrevealed ancestors. Its correlation is the product
of the edge correlations, and conditioning only shrinks each of them below
$\theta$. A $\chi^2$ recursion over the batch then bounds the cost of a
leaf by $C\theta^{4d}$, uniformly over histories. For the lower bound, call
an ancestor of height $k$ \emph{pinned} if the generated leaves give it a
posterior magnetization above $(b\theta^2)^{k/4}$. By the decay of
information below the threshold, the probability that it is ever pinned is
at most $C(b\theta^2)^{k/2}$, and Doob's maximal inequality for the posterior
magnetization makes this uniform over the rounds of an adaptive rule. Pairs
whose path avoids pinned ancestors lose information of order $\theta^{4d}$,
and a covering argument shows that a rule with $D\le b^d/2$ rounds must put
a positive fraction of the leaves in such pairs. Appendix~\ref{app:binary-below}
gives the proof.

\subsection{Above the Reconstruction Threshold}\label{sec:above}

When $b\theta^2>1$, a regular pattern of generated leaves with one leaf in
every subtree of height $s$ determines all ancestors above height $\kappa
s$, where $\kappa=1/(1-2x)$, to constant accuracy (Lemma~\ref{lem:saturation}
in Appendix~\ref{app:gauss}). Above that height, the correlation of a later
pair decays faster per edge than the edge correlation: in the Gaussian model it becomes
$1/(b\rho)$. A schedule therefore generates leaves in stages of increasing
density. Sparse stages pay for ancestors that are not yet saturated, and
dense stages pay the faster decay. For the Gaussian model this gives the
following.

\begin{theorem}[Gaussian tree]\label{thm:gauss}
There are constants $c,C>0$, depending only on $(b,\rho)$, such that the
following hold for all $h\ge1$ and $0<\eps\le1/2$.
\begin{enumerate}[label=(\roman*)]
\item If $b\rho^2<1$, then $c\min\{n,(n/\eps)^{\alpha}\}\le D_\eps\le
C\min\{n,(n/\eps)^{\alpha}\}$ with $\alpha=1/(4x)$.
\item If $b\rho^2=1$, then $c\min\{n,\sqrt{n/\eps}/h^2\}\le D_\eps\le
C\min\{n,\sqrt{n/\eps}\}$.
\item If $b\rho^2>1$, put $\alpha_{\mathrm{lo}}=1/(4(1-x))$ and
$\alpha_{\mathrm{up}}=\max\{2x/(1+2x),\,1/(4(1-x))\}$. Then
\[
 c\min\{n,(n/\eps)^{\alpha_{\mathrm{lo}}}\}\le D_\eps\le C\min\{n,\ h\,(nh^2/\eps)^{\alpha_{\mathrm{up}}}\}.
\]
\end{enumerate}
The upper bounds are attained by fixed schedules.
\end{theorem}

Both exponents in (iii) are smaller than $1/2$ for $0<x<1/2$, and they
coincide, so that the depth exponent is determined, exactly when
$b\rho^4\ge1$.

In the binary model the generated values matter, and the saturation of an
ancestor is a random event. We control it through a Bayesian backward
induction along tree paths. For $a\in[0,2)$ with $a\theta^2<1$, put
$w_a(t)=1-at^2$, $G(m)=\bigl((1-m^2)/(1-\theta^2m^2)\bigr)^2$,
$Q_a(m)=G(m)(1-a\theta^2m^2)$, and, for $|t|\le\theta$ and $0\le\tau\le1$,
\[
 \Psi_a(t,\tau)=\tfrac12\Bigl[(1+t\tau)\,Q_a\Bigl(\tfrac{t+\tau}{1+t\tau}\Bigr)+(1-t\tau)\,Q_a\Bigl(\tfrac{t-\tau}{1-t\tau}\Bigr)\Bigr].
\]
The function $\Psi_a$ is the expected squared edge factor, weighted by
$1-a\theta^2m^2$, after one ancestor receives an independent signal of
strength $\tau$. Let $\widehat\Psi_{a,t}$ be the least concave majorant
on $[0,1]$ of the function $\tau^2\mapsto\Psi_a(t,\tau)$, and let
$\lambda_a(y)=\theta^2\sup_{|t|\le\theta}\sup_{y\le\tau^2\le1}\widehat\Psi_{a,t}(\tau^2)/w_a(t)$. Put
$T=b\theta^2$ and $y^\ast_b=\lambda_s/(1+\lambda_s)$ with
$\lambda_s=(b-1)(T-1)/(b-T)$; this is the squared correlation that a node
receives from $b-1$ fully generated sibling subtrees in the Gaussian model
with $\rho=\theta$. Define
\[
 \lambda_b(\theta)=\inf\bigl\{\lambda_a(y^\ast_b):\ a\in[0,2),\ a\theta^2<1,\ \lambda_a(0)\le\theta^2\bigr\}.
\]

\begin{theorem}[Binary tree above the threshold]\label{thm:binary-above}
If $b\ge2$, $\theta\in[0.02,0.98]$ and $b\theta^2>1$, then
$b\,\lambda_b(\theta)<1$, so that
\[
 a_{\mathrm{bin}}(\theta)=\max\Bigl\{\frac{2x}{1+2x},\ \frac{\log b}{2\log(1/\lambda_b(\theta))}\Bigr\}<\frac12 ,
\]
and for every $\gamma>a_{\mathrm{bin}}(\theta)$ there is $C_\gamma$,
depending only on $(b,\theta,\gamma)$, such that, for all $h\ge1$ and
$0<\eps\le1/2$,
\[
 D_\eps\ \le\ C_\gamma\,\min\bigl\{n,\ h\,(nh^2/\eps)^{\gamma}\bigr\}.
\]
The bound is attained by a fixed schedule.
\end{theorem}

The constant $\lambda_b(\theta)$ plays the role of the squared saturated
edge correlation $1/(b\rho)^2$ of the Gaussian model. The condition
$b\lambda_b(\theta)<1$ is verified by interval arithmetic with outward
rounding (Appendix~\ref{app:binary-above}), and the theorem gives no
lower bound.

The proof of Theorem~\ref{thm:binary-above} starts from the path
structure. Conditionally on the generated leaves, the correlation of two later leaves
is a product of edge factors along their path, and each factor is a
function of the posterior magnetizations at its endpoints. The expected
product does not factorize, because the magnetizations along a path are
dependent. We order the path from the leaf upwards and treat each ancestor
as receiving the signal of its other subtrees as a Bayesian experiment.
Backward induction with the weight $w_a$ bounds the expected product by
$\prod_k\lambda_k$, and the one-step bound reduces to the concavity
inequality defining $\lambda_a$. For pairs of leaves, the two half-paths
are coupled through their common ancestor; a Doob martingale with
orthogonal increments decouples them. The schedule and the assembly are the
same as in the Gaussian model (Appendix~\ref{app:binary-above}).

\subsection{The Peak at the Reconstruction Threshold}\label{sec:peak}

For fixed $\eps$ write $a(\theta)=\limsup_{h\to\infty}\log D_\eps/\log n$
for the binary model and $a(\rho)$ for the Gaussian model.

\begin{corollary}[Peak of the depth exponent]\label{cor:peak}
In the Gaussian model, $a(b^{-1/2})=1/2$ and $a(\rho)<1/2$ for every other
$\rho\in(0,1)$; moreover $a(\rho)\to1/2$ as $\rho\to b^{-1/2}$ from either
side. In the binary model, $a(\theta)=\alpha<1/2$ for $\theta<b^{-1/2}$
with $a(\theta)\uparrow1/2$ as $\theta\uparrow b^{-1/2}$, and
$a(\theta)\le a_{\mathrm{bin}}(\theta)<1/2$ for $\max\{b^{-1/2},0.02\}<\theta\le0.98$.
\end{corollary}

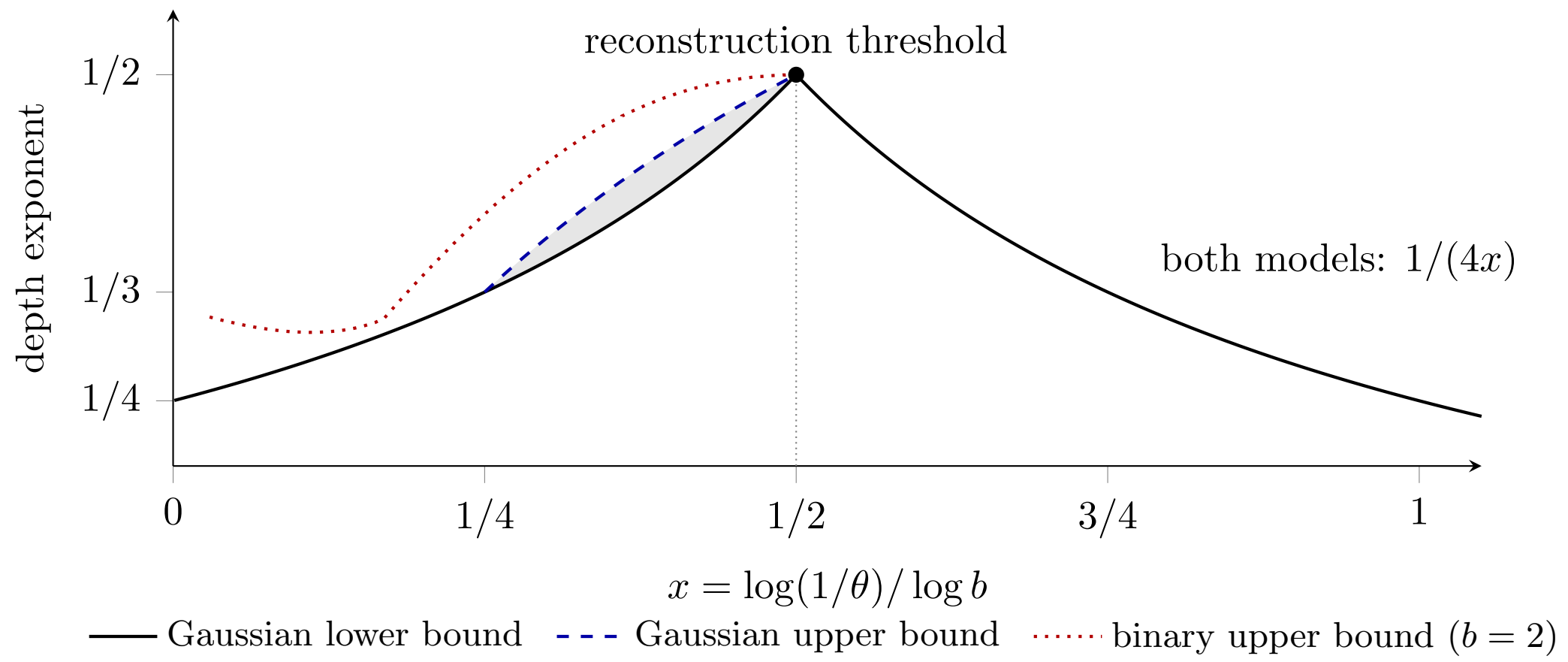
\begin{figure}[t]
\centering
\begin{tikzpicture}
\begin{axis}[
 width=.86\linewidth,height=5.6cm,
 xmin=0,xmax=1.05,ymin=0.2,ymax=.55,
 axis lines=left, tick align=outside,
 xlabel={$x=\log(1/\theta)/\log b$},ylabel={depth exponent},
 xtick={0,.25,.5,.75,1},xticklabels={$0$,$1/4$,$1/2$,$3/4$,$1$},
 ytick={.25,.333333,.5},yticklabels={$1/4$,$1/3$,$1/2$},
 label style={font=\small},tick label style={font=\small},
 legend style={font=\footnotesize,at={(0.5,-0.3)},anchor=north,draw=none,fill=none,legend columns=3,/tikz/every even column/.append style={column sep=6pt}},
 clip=false]
\addplot[name path=upper,draw=none,domain=.25:.5,samples=100,forget plot] {2*x/(1+2*x)};
\addplot[name path=lower,draw=none,domain=.25:.5,samples=100,forget plot] {1/(4*(1-x))};
\addplot[gray!20,forget plot] fill between[of=upper and lower];
\addplot[black,thick,domain=.001:.5,samples=80] {1/(4*(1-x))};
\addlegendentry{Gaussian lower bound}
\addplot[blue!65!black,thick,dashed,domain=.25:.5,samples=60] {2*x/(1+2*x)};
\addlegendentry{Gaussian upper bound}
\addplot[red!70!black,thick,dotted] table[x=x,y=a] {binary_upper_b2.dat};
\addlegendentry{binary upper bound ($b=2$)}
\addplot[black,thick,domain=.5:1.05,samples=100,forget plot] {1/(4*x)};
\addplot[gray,densely dotted,forget plot] coordinates {(.5,.2)(.5,.5)};
\addplot[only marks,mark=*,mark size=1.8pt,forget plot] coordinates {(.5,.5)};
\node[anchor=south,font=\small] at (axis cs:.5,.505) {reconstruction threshold};
\node[anchor=south west,font=\small] at (axis cs:.78,.33) {both models: $1/(4x)$};
\end{axis}
\end{tikzpicture}
\caption{Bounds on the exponent of $n$ in the leaf depth at fixed
accuracy. Right of the threshold ($x>1/2$) both models have the exponent
$1/(4x)$. Left of it, the Gaussian exponent lies in the shaded band and is
determined outside it; the binary upper bound ($b=2$, $\theta\le0.98$) is
computed from $\lambda_b(\theta)$ in Theorem~\ref{thm:binary-above}. No
lower bound is shown for the binary model above the threshold.}
\label{fig:phase}
\end{figure}

Figure~\ref{fig:phase} shows the bounds. The peak has a direct reading for
plan-then-fill decoding. A generated plan replaces a power of $n$ by
$\log n$ rounds (Section~\ref{sec:plan}), and the power is largest at the
threshold. Weakly correlated data need few rounds without a plan. Strongly
correlated data need fewer rounds than at the threshold because the leaves
generated in sparse stages carry an implicit plan. The gain from an
explicit plan is largest in between.

The upper bounds share one schedule: generate leaves that are far apart
in the tree in the same round, and generate sparse leaves before dense
ones. The first part keeps every batch nearly independent below the
threshold. The second part lets later batches condition on ancestors that
earlier rounds have determined. Rules that select positions independently
of the tree, such as uniformly random batches, or that select the most
confident positions first, do not keep the leaves of a batch apart in the tree.
Section~\ref{sec:experiments} measures how much this costs.

\section{Generating the Plan}\label{sec:plan}

This section asks how many rounds a decoder needs when it may generate the
internal nodes, and how fine the plan must be. Section~\ref{sec:full}
shows that the full depth is logarithmic, and Section~\ref{sec:granularity}
treats plans that stop above the leaves.

\subsection{Full Access to the Latent Variables}\label{sec:full}

\begin{theorem}[Generating the plan]\label{thm:plan}
In both models, $D^{\mathrm{all}}_0\le h+1$. In the binary model, every
deterministic value-adaptive rule on all nodes satisfies
\[
 D^{\mathrm{all}}_\eps\ \ge\ \min\Bigl\{h+1,\ \frac{\log(c_0 n/\eps)}{\log b+4\log(1/\beta_\ast)}\Bigr\},
\]
where $\beta_\ast=\tfrac12[\tanh((b+1)J)-\tanh((b-1)J)]$,
$J=\operatorname{artanh}\theta$, and
$c_0=\tfrac12\bigl(1-\tanh^2((b+1)J)\bigr)^2$. In particular, every such
rule $A$ that uses at most $R\le h$ rounds has
$\KL(P\,\|\,Q_A)\ge c_0\,n\,(\beta_\ast^4/b)^R$.
\end{theorem}

The upper bound generates the tree one level per round: given a level,
the nodes of the next level are conditionally independent, so every round
has zero total correlation. Together with
Theorems~\ref{thm:binary-below}--\ref{thm:binary-above}, the theorem
separates the two decoders. Below the threshold the leaf depth is of order
$\min\{n,(n/\eps)^{\alpha}\}$, and in the binary model the full depth is
of order $\log n$; both orders are attained. The lower bound concerns the joint law of all nodes,
as appropriate when the plan must itself be correct. It does not transfer
to the leaf law, since the leaf divergence can be smaller than the joint
one.

The lower bound of Theorem~\ref{thm:plan} is a tree-depth argument. If all nodes of a subtree of
height $H$ are generated in at most $H$ rounds, some round contains two
nodes of the subtree whose connecting path contains no earlier generated
node: the first round that touches the subtree generates at most one of
its nodes, and removing that node leaves a complete subtree of height
$H-1$ to be generated in fewer rounds. Such an unseparated pair at distance
$\ell$ loses mutual information at least $c_0\beta_\ast^{2\ell}$ whatever
else has been generated, because every message along the path is bounded.
Counting disjoint subtrees gives the bound (Appendix~\ref{app:plan}).

\subsection{Granularity of the Plan}\label{sec:granularity}

A practical plan does not reach down to single tokens. Suppose that the
decoder may generate the leaves and the nodes of height at least $s$. It
may not generate the nodes strictly between, so every finest plan node
covers $m=b^s$ leaves.

\begin{theorem}[Granularity]\label{prop:granularity}
In the binary model with $b\theta^2<1$ there are constants $c,C>0$,
depending only on $(b,\theta)$, such that, for all $h\ge s\ge1$ and
$0<\eps\le1/2$,
\[
 c\,\bigl[(h-s+1)+\min\{m,(n/\eps)^{\alpha}\}\bigr]\ \le\ D_\eps^{(s)}\ \le\ (h-s+1)+C\min\{m,(n/\eps)^{\alpha}\},
\]
where $D^{(s)}_\eps$ is the depth for the joint law of the nodes of height
at least $s$ and the leaves. The lower bound holds for every deterministic
value-adaptive rule, which may generate plan nodes and leaves in any order.
\end{theorem}

In particular, for fixed $b$, $\theta$ and $\eps$, the depth is $O(\log
n)$ if and only if $m=O(\log n)$, and it is of order $m$ when $\log n\ll
m\ll(n/\eps)^{\alpha}$.

The upper bound generates the plan level by level and then runs the
schedule of Theorem~\ref{thm:binary-below} inside every finest subtree;
given their roots, the subtrees are independent. The lower bound has two
parts. The term $h-s+1$ comes from the argument for
Theorem~\ref{thm:plan}, applied to the plan nodes: generated leaves never
lie on the path between two plan nodes. The term
$\min\{m,(n/\eps)^{\alpha}\}$ extends the argument for
Theorem~\ref{thm:binary-below}: a generated plan node of height $s$ pins
the ancestors below it only through a path of length $s-k$, so the pinning
probability of an ancestor of height $k<s$ gains a term $\theta^{s-k}$
(Appendix~\ref{app:plan}).

The theorem gives a rule for the size of plan units. A plan reaches
logarithmic depth exactly when its units are of logarithmic size, and a
plan whose finest units cover more than $(n/\eps)^{\alpha}$ tokens reduces
the depth by at most a constant factor. More generally, a plan helps only through the
conditional independence it creates: given the plan, the remaining
dependence among the tokens of different units must vanish. This is the
property that planned diffusion asks its plan to have when it partitions a
response into chunks meant to be independent \citep{israel2026}. On a
tree, the condition can be checked: units are independent given their
roots, and the depth inside a unit is again governed by
Theorem~\ref{thm:binary-below}.

\section{Experiments}\label{sec:experiments}

This section computes the depth of parallel decoders on trees with
$b=2$, in the binary model and, for larger trees, in the Gaussian model,
and compares it with Sections~\ref{sec:leaf} and~\ref{sec:plan}.
Section~\ref{sec:exp-setting} describes the setting and the four rules.
Section~\ref{sec:exp-below} checks the leaf depth below the threshold
against Theorem~\ref{thm:binary-below}. Section~\ref{sec:exp-above}
compares staged schedules with the stride schedule above the threshold,
in the binary and in the Gaussian model. Section~\ref{sec:exp-rules}
measures rules that ignore the tree, and Section~\ref{sec:exp-plan}
measures plans of different granularity.

\subsection{Setting and Rules}\label{sec:exp-setting}

The conditionals are exact, so the measured divergence is the conditional
total correlation of the batches and contains no model error. We take
$\theta\in\{0.40,0.50,\dots,0.90\}$, which covers both sides of
the threshold $\theta=2^{-1/2}$, and $n=2^h$ with $h=6,\dots,12$, and
$h\le13$ for the multistage schedules. The
depth $D_\eps$ is read off the divergence at $D=1,2,4,\dots$, and its
exponent is the least-squares slope of $\log D_\eps$ against $\log n$ over
$h=9,\dots,12$. Appendix~\ref{app:exp} describes how the divergence is
computed.

We compare four rules. The stride schedule with $D=2^k$ rounds generates
in each round one leaf of every subtree of height $k$, and orders the
rounds from sparse to dense; it attains the upper bound of
Theorem~\ref{thm:binary-below}. The multistage schedules of
Appendix~\ref{app:gauss-above} let the height of these subtrees grow
from one stage to the next, as in the proofs of
Theorems~\ref{thm:gauss} and~\ref{thm:binary-above}. We choose the number
of rounds of every stage to minimize the total, and this family includes
the stride schedule as a special case. Uniformly random batches are the default of
masked diffusion samplers, and the confidence rule generates the leaves
with the largest posterior $|\E[X_i\mid\text{revealed}]|$ first.

\subsection{Leaf Depth below the Threshold}\label{sec:exp-below}

Table~\ref{tab:exp-below} compares the stride schedule with
Theorem~\ref{thm:binary-below}, which predicts $D_\eps\asymp
(n/\eps)^{\alpha}$. For $\theta\le0.65$ the measured exponent differs from
$\alpha$ by at most $0.07$, and the ratio $D_{0.03}/D_{0.1}$ agrees with
$(10/3)^{\alpha}$. Closer to the threshold the measured exponent is
smaller than $\alpha$ at these sizes.

\begin{table}[t]
\centering
\small
\begin{tabular}{@{}lccccc@{}}
\toprule
$\theta$ & $\alpha$ & measured exponent & $D_{0.1}$ at $n=4096$ & $D_{0.03}/D_{0.1}$ & $(10/3)^{\alpha}$\\
\midrule
0.40 & 0.189 & 0.180 & 2.8 & 1.25 & 1.26\\
0.50 & 0.250 & 0.245 & 4.5 & 1.38 & 1.35\\
0.60 & 0.339 & 0.327 & 9.0 & 1.43 & 1.50\\
0.65 & 0.402 & 0.336 & 14.5 & 1.67 & 1.62\\
0.70 & 0.486 & 0.385 & 25.6 & 1.64 & 1.80\\
\bottomrule
\end{tabular}
\caption{Stride schedule below the threshold, $\eps=0.1$. The exponent is
the slope of $\log D_\eps$ against $\log n$ for $n=2^9,\dots,2^{12}$, and
$\alpha=\log2/(4\log(1/\theta))$.}
\label{tab:exp-below}
\end{table}

\subsection{Staged Schedules above the Threshold}\label{sec:exp-above}

\EXPABOVE

\begin{table}[t]
\centering
\small
\begin{tabular}{@{}lccccc@{}}
\toprule
& & \multicolumn{2}{c}{measured exponent} & \multicolumn{2}{c}{$D_{0.1}$ at $n=8192$}\\
\cmidrule(lr){3-4}\cmidrule(l){5-6}
$\theta$ & $a_{\mathrm{bin}}$ & stride & multistage & stride & multistage\\
\midrule
\EXPABOVEROWS
\bottomrule
\end{tabular}
\caption{Stride and multistage schedules above the threshold,
$\eps=0.1$. The exponent is the slope of $\log D_\eps$ against $\log n$
for $n=2^8,\dots,2^{13}$, with its standard error in parentheses, and
$a_{\mathrm{bin}}$ is the upper exponent of
Theorem~\ref{thm:binary-above}.}
\label{tab:exp-above}
\end{table}

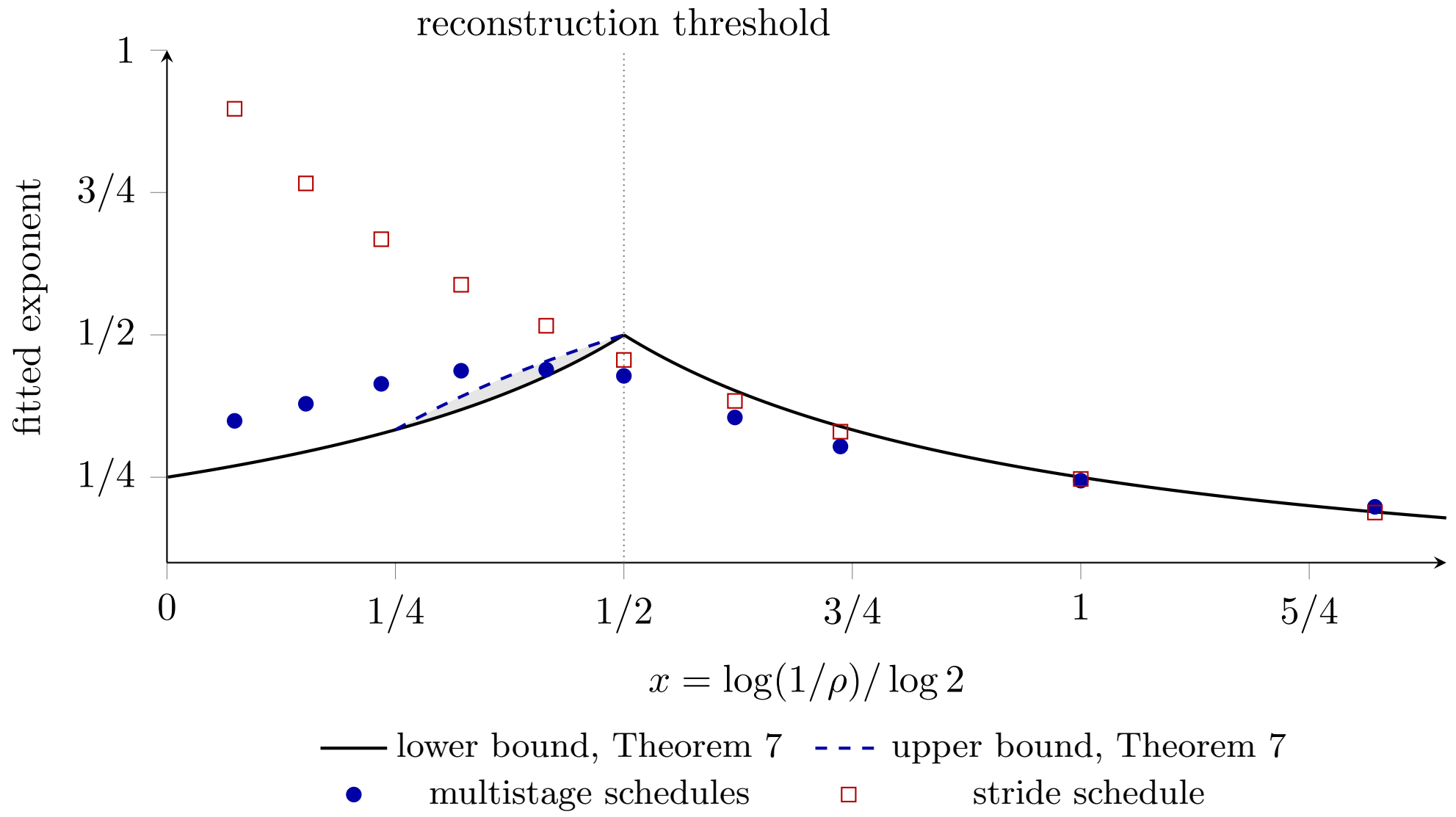
\begin{figure}[t]
\centering
\begin{tikzpicture}
\begin{axis}[
 width=.86\linewidth,height=6.2cm,
 xmin=0,xmax=1.4,ymin=0.1,ymax=1,
 axis lines=left, tick align=outside,
 xlabel={$x=\log(1/\rho)/\log 2$},ylabel={fitted exponent},
 xtick={0,.25,.5,.75,1,1.25},xticklabels={$0$,$1/4$,$1/2$,$3/4$,$1$,$5/4$},
 ytick={.25,.5,.75,1},yticklabels={$1/4$,$1/2$,$3/4$,$1$},
 label style={font=\small},tick label style={font=\small},
 legend style={font=\footnotesize,at={(0.5,-0.3)},anchor=north,draw=none,fill=none,legend columns=2,/tikz/every even column/.append style={column sep=6pt}},
 clip=false]
\addplot[name path=upper,draw=none,domain=.25:.5,samples=100,forget plot] {2*x/(1+2*x)};
\addplot[name path=lower,draw=none,domain=.25:.5,samples=100,forget plot] {1/(4*(1-x))};
\addplot[gray!20,forget plot] fill between[of=upper and lower];
\addplot[black,thick,domain=.001:.5,samples=80] {1/(4*(1-x))};
\addlegendentry{lower bound, Theorem~\ref{thm:gauss}}
\addplot[blue!65!black,thick,dashed,domain=.25:.5,samples=60] {2*x/(1+2*x)};
\addlegendentry{upper bound, Theorem~\ref{thm:gauss}}
\addplot[black,thick,domain=.5:1.4,samples=100,forget plot] {1/(4*x)};
\addplot[gray,densely dotted,forget plot] coordinates {(.5,.1)(.5,1)};
\addplot[only marks,mark=*,mark size=1.8pt,color=blue!65!black] coordinates {
 (1.3219,.198)(1,.244)(.7370,.304)(.6215,.355)(.5,.428)
 (.4150,.439)(.3219,.437)(.2345,.414)(.1520,.379)(.0740,.349)};
\addlegendentry{multistage schedules}
\addplot[only marks,mark=square,mark size=1.8pt,color=red!70!black] coordinates {
 (1.3219,.188)(1,.247)(.7370,.330)(.6215,.384)(.5,.456)
 (.4150,.516)(.3219,.588)(.2345,.668)(.1520,.766)(.0740,.897)};
\addlegendentry{stride schedule}
\node[anchor=south,font=\small] at (axis cs:.5,1) {reconstruction threshold};
\end{axis}
\end{tikzpicture}
\caption{Gaussian model, $b=2$, $\eps=0.1$: least-squares slope of $\log
D_\eps$ against $\log n$ for $n=2^{14},\dots,2^{20}$, with divergences
computed exactly, and the
exponents of Theorem~\ref{thm:gauss}. The upper bound coincides with the
lower bound outside the shaded band. The threshold is at $x=1/2$, and
$x$ decreases as $\rho$ increases.}
\label{fig:exp-gauss}
\end{figure}

In the Gaussian model the conditional covariances do not depend on the
generated values, so the divergence of a schedule is a deterministic
function of its batches and is computed without sampling. We compute the
stage costs of the multistage schedules for
$\rho\in\{0.40,0.50,0.60,0.65,2^{-1/2},0.75,0.80,0.85,0.90,0.95\}$ and
$n=2^h$ with $h=8,\dots,20$, and choose the allocation as in the binary
model; Appendix~\ref{app:exp} gives the method. Figure~\ref{fig:exp-gauss}
shows the exponents fitted over $h=14,\dots,20$ at $\eps=0.1$. At
$\eps=0.01$ the exponents of the multistage schedules change by at most
$0.02$, and those of the stride schedule by at most $0.01$ below the
threshold and by up to $0.07$ above it.

Below the threshold the exponent of the stride schedule is within $0.02$
of $\alpha=1/(4x)$. The multistage schedules change the depth at
$n=2^{20}$ by at most $20\%$ there, and their exponents differ from those
of the stride schedule by at most $0.03$. Above the threshold the two
schedules separate. The exponent of the stride schedule keeps increasing
with $\rho$ and reaches $0.90$ at $\rho=0.95$. The exponent of the
multistage schedules is largest, between $0.43$ and $0.44$, for $\rho$
between $2^{-1/2}$ and $0.8$, and decreases to $0.35$ at $\rho=0.95$. At
$\rho=0.95$ and $n=2^{20}$ the selected multistage schedule reaches
divergence $0.095$ in $348$ rounds, and the interpolated depth of the
stride schedule is about $2.5\cdot10^{5}$.

These exponents agree with Theorem~\ref{thm:gauss} up to its logarithmic
factors: at the threshold and for $\rho\ge0.85$ they lie between the
local slopes of the two bounds (Appendix~\ref{app:exp}). The location of
the peak is not resolved at these sizes: for $\rho$ between $2^{-1/2}$
and $0.8$ the measured exponents differ by at most $0.011$.

\subsection{Rules That Ignore the Tree}\label{sec:exp-rules}

Table~\ref{tab:exp-rules} lists the depth of the four rules at $n=4096$.
Uniformly random batches need $0.07n$ rounds at $\theta=0.4$ and more
than $n/2$ rounds for $\theta\ge0.75$. Where their depth stays below
$n/2$, its measured exponent lies between $0.93$ and $1.15$. The figures
are averages over seeds, which bound the divergence of the randomized
rule from above (Remark~\ref{rem:randomized}). The confidence rule is
worse: at $D=n/2$ its divergence is between $9$ and $82$ nats, so it
needs more than $n/2$ rounds at every $\theta$.

\begin{table}[t]
\centering
\small
\begin{tabular}{@{}lcccc@{}}
\toprule
$\theta$ & stride & multistage & random batches & confidence\\
\midrule
\EXPRULESROWS
\bottomrule
\end{tabular}
\caption{Depth $D_{0.1}$ at $n=4096$. Entries $>2048$ mean that the
divergence at $D=n/2$ still exceeds $\eps$. The stride and multistage
entries come from independent samples; below the threshold the two
schedules coincide up to sampling error.}
\label{tab:exp-rules}
\end{table}

\subsection{Plans of Different Granularity}\label{sec:exp-plan}

Table~\ref{tab:exp-plan} generates the internal nodes of depth at most
$h-g$ level by level and then the leaves by the stride schedule, so that
every finest plan node covers $m=2^g$ leaves. The plan levels cost
$h-g+1$ rounds and no divergence. \EXPPLAN

\begin{table}[t]
\centering
\small
\begin{tabular}{@{}lccccccc@{}}
\toprule
& no plan & \multicolumn{6}{c}{plan with units of $m=2^g$ leaves}\\
\cmidrule(l){3-8}
$\theta$ & & $m=1024$ & $256$ & $64$ & $16$ & $4$ & $2$\\
\midrule
\EXPPLANROWS
\bottomrule
\end{tabular}
\caption{Total rounds at $n=4096$ and $\eps=0.1$: plan levels plus leaf
rounds. The column without a plan uses the multistage schedule.}
\label{tab:exp-plan}
\end{table}

\section{Conclusion}\label{sec:conclusion}

We asked when a plan saves parallel rounds, how many it saves, and how
fine it must be. On broadcast trees, generated leaves never separate the
remaining ones, so without a plan every deterministic value-adaptive rule
needs of order $\min\{n,(n/\eps)^{\alpha}\}$ rounds below the
reconstruction threshold. A plan reduces this to $h+1$ rounds. The saving
is largest near the threshold, where $\alpha$ reaches $1/2$. Weakly
correlated tokens need few rounds even without a plan. Strongly correlated
tokens carry an implicit plan, and staged schedules keep the exponent
below $1/2$ above the threshold, in the binary model within the verified
range. In the binary model, a plan whose units cover $m$ tokens needs of
order $\log_b(n/m)+\min\{m,(n/\eps)^{\alpha}\}$ rounds, so it gives
logarithmic depth exactly when its units have logarithmic size. Without a
plan, tokens far apart in the tree should share a round, and sparse tokens
should precede dense ones. Exact computations confirm the exponents below
the threshold.

The lower bounds hold for learned conditionals but concern deterministic
rules on regular trees with a known channel. Randomized rules, a binary
lower bound above the threshold and trees of unknown structure are left
open.

\acks{The authors used Claude (Anthropic) and ChatGPT (OpenAI) for language
polishing, typesetting, proof checking, and code writing. The authors
checked all content and take full responsibility for it. The authors
declare no competing interests.}

\appendix
\makeatletter
\renewcommand\theHsection{\Alph{section}}
\makeatother
\section{Cost Identity and Learned Conditionals}\label{app:cost}

This appendix proves Propositions~\ref{prop:cost} and~\ref{prop:learned}
and records two elementary facts about total correlation used later. The
identity is the decomposition of \citet{benhamu2025} for deterministic
rules. We prove it in the present form, which lets the rule choose its
batches from the predictions and compares it with the rule that uses the
same batches with exact conditionals. We write the proof for finite
alphabets; for the Gaussian model the same argument applies to densities.

\begin{proof}[Proof of Propositions~\ref{prop:cost} and~\ref{prop:learned}]
Fix a deterministic rule $A$ and conditionals $\widehat P_i(\cdot\mid
x_R)$, which may equal the exact ones. For $x$ with $P(x)>0$ the rule
produces a unique trajectory $(R_t(x),S_t(x))_{t\le D}$, and every visited
context has $P(x_{R_t})\ge P(x)>0$. The chain rule along the trajectory
gives
\[
 P(x)=\prod_t P\bigl(x_{S_t}\mid x_{R_t}\bigr),\qquad
 \widehat Q_A(x)=\prod_t\prod_{i\in S_t}\widehat P_i\bigl(x_i\mid x_{R_t}\bigr).
\]
Hence
\[
 \log\frac{P(x)}{\widehat Q_A(x)}=\sum_t\Bigl[\log\frac{P(x_{S_t}\mid x_{R_t})}{\prod_{i\in S_t}P(x_i\mid x_{R_t})}
 +\sum_{i\in S_t}\log\frac{P(x_i\mid x_{R_t})}{\widehat P_i(x_i\mid x_{R_t})}\Bigr].
\]
The batch $S_t$ and the context $R_t$ are functions of $x_{R_t}$ alone,
because every earlier choice used only values in $x_{R_t}$. The event that
the rule reaches the state $(R_t,x_{R_t})$ is therefore the cylinder event
$\{X_{R_t}=x_{R_t}\}$, and conditioning on the state introduces no
selection bias. Taking the conditional expectation of the bracket given
the state turns the first term into $\TC(X_{S_t}\mid x_{R_t})$ and the
second into $\sum_{i\in S_t}\KL(P(\cdot\mid x_{R_t})\,\|\,\widehat
P_i(\cdot\mid x_{R_t}))$. Summing over $t$ and taking expectations proves
the identity of Proposition~\ref{prop:learned}; with exact conditionals the
second term vanishes, which is Proposition~\ref{prop:cost}.

For the comparison, let $A'$ be the rule with exact conditionals whose
batch at every state is the batch that $A$ chooses there. If $A$ uses the
predictions to choose its batches, the choice is still a deterministic
function of the generated values, so $A'$ is a deterministic
value-adaptive rule with the same round count, and its trajectories under
$P$ are those of $A$. The identity then gives
$\KL(P\,\|\,\widehat Q_A)\ge\KL(P\,\|\,Q_{A'})$. If a rule with learned
conditionals meets the budget $\eps$ in $D$ rounds, so does $A'$; hence
every lower bound on $D_\eps$ also bounds the rounds of decoders with
learned conditionals.
\end{proof}

\begin{lemma}[Total correlation of a batch]\label{lem:tc}
For disjoint $S,R$ and a context $x_R$:
\begin{enumerate}[label=(\roman*)]
\item if $S=S_1\cup\dots\cup S_k$ is a partition, then
$\TC(X_S\mid x_R)\ge\sum_j\TC(X_{S_j}\mid x_R)$;
\item if $S$ is ordered as $s_1,s_2,\dots$, then
$\TC(X_S\mid x_R)=\sum_{i\ge2}I(X_{s_i};X_{s_1},\dots,X_{s_{i-1}}\mid x_R)$.
\end{enumerate}
\end{lemma}

Both parts follow from the chain rule for entropy. Part (ii) and the
monotonicity of mutual information in its second argument give the
collision count of Lemma~\ref{lem:collision}.

\section{Binary Tree Below the Threshold}\label{app:binary-below}

This appendix proves Theorem~\ref{thm:binary-below}. Appendix~\ref{app:tools}
collects facts about binary Markov chains and $\chi^2$ information,
Appendix~\ref{app:below-upper} proves the upper bound, and
Appendix~\ref{app:below-lower} the lower bound. We write
$J=\operatorname{artanh}\theta$, $u=\log(1/\theta)$, $L=\log b$, and
$I_{\chi^2}(U;W)=\chi^2(P_{UW}\,\|\,P_U\otimes P_W)$, so that
$I\le I_{\chi^2}$. Conditionally on any set of generated nodes, the tree
model is again an Ising model on the tree with external fields at the
neighbors of generated nodes.

\subsection{Tools}\label{app:tools}

For $h\ge0$ let $s(h)=\sinh2J/(\cosh2h+\cosh2J)$; this is the slope of
$\E[Z'\mid Z]$ for an edge whose receiving end $Z'$ has external field $h$
from its other neighbors. The function $r(h)=s(h)/\theta\in(0,1]$ decreases
in $|h|$.

\begin{lemma}\label{lem:chain}
\begin{enumerate}[label=(\roman*)]
\item (Decomposition.) If $U-V-W$ is a Markov chain with $U,V$ binary, then
$I_{\chi^2}(U;W)=\rho^2(U,V)\,I_{\chi^2}(V;W)$.
\item (Edge contraction.) Conditionally on any generated values, two
adjacent nodes with external fields $h,h'$ from their other neighbors have
squared correlation $\theta^2r(h)r(h')\le\theta^2$. Along a path of
unrevealed nodes, the squared correlation of the endpoints is the product
of the squared edge correlations.
\end{enumerate}
\end{lemma}

\begin{proof}
(i) $\E[U\mid W]=\E[\E[U\mid V]\mid W]=a+s\,\E[V\mid W]$, so $\Var\E[U\mid
W]=s^2\Var\E[V\mid W]$. For binary variables $I_{\chi^2}(U;W)=\Var\E[U\mid
W]/\Var U$, and $s^2\Var V/\Var U=\rho^2(U,V)$.

(ii) The joint law of the two nodes is proportional to $\exp(Jzz'+hz+h'z')$,
whose squared correlation is $s(h)s(h')=\theta^2r(h)r(h')$. A binary
inhomogeneous Markov chain has $\rho(U,W)=\rho(U,V)\rho(V,W)$ by the
argument of~(i).
\end{proof}

\begin{lemma}[Decay of information]\label{lem:decay}
For every set $R$ of leaves and every node $v$,
\[
 I(Z_v;X_R)\le\sum_{\ell\in R}\theta^{2\operatorname{dist}(v,\ell)},\qquad
 \E\bigl[\E[Z_v\mid X_R]^2\bigr]\le2\sum_{\ell\in R}\theta^{2\operatorname{dist}(v,\ell)}.
\]
If $b\theta^4<1$ and $v$ has height $k$, the right-hand side is at most
$C_x(b\theta^2)^k$ with $C_x=2/(1-b\theta^4)$. The same bounds hold when
$R$ also contains generated internal nodes, with the distance to each of
them.
\end{lemma}

\begin{proof}
Order $R$ as $\ell_1,\ell_2,\dots$. By the chain rule,
$I(Z_v;X_R)=\sum_i\E\,I(Z_v;X_{\ell_i}\mid x_{\ell_1},\dots,x_{\ell_{i-1}})$.
Given the earlier values, $Z_v$ and $X_{\ell_i}$ are binary and connected by
a path of unrevealed nodes, which is a binary Markov chain; by
Lemma~\ref{lem:chain}(ii) each term is at most
$\rho^2\le\theta^{2\operatorname{dist}(v,\ell_i)}$. Since $Z_v$ is uniform,
Pinsker's inequality gives $\KL\ge M^2/2$ for the posterior with mean $M$,
which proves the second bound. For the third, the subtree of $v$ contributes
$b^k\theta^{2k}$, and the leaves that branch off at the $i$th ancestor of
$v$ contribute at most $(b-1)b^{k+i-1}\theta^{2(k+2i)}$; the sum over $i$ is
geometric because $b\theta^4<1$.
\end{proof}

\subsection{Upper Bound}\label{app:below-upper}

Consider the fixed schedule that, in round $t\le b^d$, generates the $t$th
leaf of every subtree of height $d$.

\begin{lemma}[Sparse batches]\label{lem:sparse}
Let $b\theta^4<1$. In every round of this schedule, for every history and
every leaf $i$ of the batch $S$,
\[
 I\bigl(X_i;X_{S\setminus i}\mid x_R\bigr)\le\frac{(b-1)\theta^{4d+4}}{1-b\theta^4}.
\]
The same bound holds when some internal nodes have been generated.
\end{lemma}

\begin{proof}
Order the leaves of $S\setminus i$ as $j_1,j_2,\dots$. By the chain rule,
\[
 I\bigl(X_i;X_{S\setminus i}\mid x_R\bigr)=\sum_k I\bigl(X_i;X_{j_k}\mid X_{j_1},\dots,X_{j_{k-1}},x_R\bigr).
\]
Fix the values of $x_R$ and $X_{j_1},\dots,X_{j_{k-1}}$. The conditional
law is again a tree model in which these values act as external fields,
and $(X_i,X_{j_k})$ is a binary pair. Its mutual information is at most
its $\chi^2$ information, which is the squared conditional correlation. If
the path from $i$ to $j_k$ contains a generated node, the two leaves are
conditionally independent and the term vanishes. Otherwise
Lemma~\ref{lem:chain}(ii) bounds the term by
$\theta^{2\operatorname{dist}(i,j_k)}$. A leaf whose nearest common
ancestor with $i$ has height $d+t$, $t\ge1$, is at distance $2(d+t)$, and
the batch has at most $(b-1)b^{t-1}$ such leaves, one in each subtree of
height $d$. Hence
\[
 I\bigl(X_i;X_{S\setminus i}\mid x_R\bigr)\le\sum_{t\ge1}(b-1)b^{t-1}\theta^{4(d+t)}
 =\frac{(b-1)\theta^{4d+4}}{1-b\theta^4}.
\]
\end{proof}

\begin{proof}[Proof of the upper bound in Theorem~\ref{thm:binary-below}]
Below the threshold $b\theta^4<\theta^2<1$. By Lemma~\ref{lem:tc}(ii) and
the monotonicity of mutual information, the total correlation of a batch is
at most the sum over its leaves of $I(X_i;X_{S\setminus i}\mid x_R)$. By
Lemma~\ref{lem:sparse}, each round costs at most $|S|\,C\theta^{4d}$ with
$C=(b-1)\theta^4/(1-b\theta^4)$, and the schedule has divergence at most
$Cn\theta^{4d}$. Choosing the smallest $d$ with $Cn\theta^{4d}\le\eps$
gives $D_\eps\le b^d\le b(Cn/\eps)^{\alpha}$, and $D_\eps\le n$ always.
\end{proof}

\subsection{Lower Bound}\label{app:below-lower}

\begin{lemma}[Collision count]\label{lem:collision}
Let the leaves be partitioned into groups $\mathcal G$, and let
$\mathcal H$ be a set of pairs (state, leaf). Suppose that
$I(X_a;X_c\mid x_R)\ge\mu$ for every state $(R,x_R)$ and every two leaves
$a,c$ of one group that both form a pair in $\mathcal H$ with this state.
Then every rule with $D$ rounds satisfies
\[
 \KL(P\,\|\,Q_A)\ge\mu\bigl(\E N_{\mathcal H}-D|\mathcal G|\bigr),
\]
where
$N_{\mathcal H}$ counts the leaves generated at a state with which they
form a pair in $\mathcal H$.
\end{lemma}

\begin{proof}
Fix a round with state $(R,x_R)$ and batch $S$, and let $S'\subseteq S$ be
the leaves that form a pair in $\mathcal H$ with this state; $S'$ is a
function of the state. By Lemma~\ref{lem:tc}(i),
$\TC(X_S\mid x_R)\ge\TC(X_{S'}\mid x_R)$. Order $S'$ so that, within each
group, its first leaf comes first. By Lemma~\ref{lem:tc}(ii) and
monotonicity of mutual information, every leaf of $S'$ that is not the
first of its group contributes at least its mutual information with that
first leaf. The batch touches at most $|\mathcal G|$ groups, so
$\TC(X_S\mid x_R)\ge\mu(|S'|-|\mathcal G|)$. Summing over the at most $D$
rounds and taking expectations gives the claim.
\end{proof}

The lemma is the star version of the disjoint-pair bound in
Section~\ref{sec:counting}. The pairwise information is controlled through
the posterior magnetizations. Let $\mathcal F_t$ be the information
generated before round $t$, and $M_v(t)=\E[Z_v\mid\mathcal F_t]$.

\begin{lemma}[Pairs and magnetizations]\label{lem:pairmag}
There is $K$ depending only on $(b,\theta)$ such that, for unrevealed
leaves $a,c$ with nearest common ancestor of height $d$ and path
$a=v_0,\dots,v_{2d}=c$,
\[
 I(X_a;X_c\mid\mathcal F_t)\ge\tfrac12(1-\theta^2)^2\theta^{4d}
 \exp\Bigl(-K\sum_{i<2d}\bigl(M_{v_i}(t)^2+M_{v_{i+1}}(t)^2\bigr)\Bigr).
\]
\end{lemma}

\begin{proof}
By Lemma~\ref{lem:chain}(ii) the squared correlation of $X_a,X_c$ is
$\prod_i\theta^2r(h_i)r(h_i')$ over the edges of the path. The exclusive
fields of an edge satisfy $M=\tanh(h+f(h'))$ and $M'=\tanh(h'+f(h))$ with
$|f(y)|\le\theta|y|$, hence $|h|+|h'|\le(|\operatorname{artanh}M|+
|\operatorname{artanh}M'|)/(1-\theta)$. All fields are at most $(b+1)J$, so
$|M|\le M_{\max}=\tanh((b+1)J)$ and $|\operatorname{artanh}M|\le|M|/
(1-M_{\max}^2)$. Further $-\log r(h)\le(\cosh2h-1)/(1+\cosh2J)\le\kappa_J
h^2$ with $\kappa_J=2\cosh(2(b+1)J)/(1+\cosh2J)$. Combining these bounds and
using $\Var X_a\ge1-\theta^2$ and $I\ge\Cov^2/2$ (Pinsker) gives the claim
with $K=2\kappa_J/((1-\theta)^2(1-M_{\max}^2)^2)$.
\end{proof}

\begin{proof}[Proof of the lower bound in Theorem~\ref{thm:binary-below}]
For every node $v$, $M_v(t)$ is a martingale in $t$ whatever the rule does,
and its final value is $\E[Z_v\mid X]$. By Doob's maximal inequality and
Lemma~\ref{lem:decay},
\[
 \Pr\Bigl(\sup_t M_v(t)^2\ge\lambda^2\Bigr)\le\frac{C_x(b\theta^2)^k}{\lambda^2}
\]
for $v$ of height $k$. Call $v$ \emph{pinned at time $t$} if
$M_v(t)^2\ge(b\theta^2)^{k/2}$, and \emph{ever pinned} if this happens for
some $t$; then $\Pr(v\text{ ever pinned})\le C_x(b\theta^2)^{k/2}$. Call a
leaf \emph{covered at time $t$} if it has an ancestor of height in
$[k_0,j]$ that is pinned at time $t$. A leaf covered at the time it is
generated has an ever-pinned ancestor, so the expected number of such
leaves is at most $nC_x(b\theta^2)^{k_0/2}/(1-(b\theta^2)^{1/2})$, which is
at most $n/4$ for a suitable $k_0$ depending only on $(b,\theta)$.

Take as groups the subtrees of height $j$, and let $\mathcal H$ consist of
the states at which the leaf to be generated is not covered; whether a leaf
is covered at time $t$ depends only on $\mathcal F_t$. For two leaves of
one group that are both uncovered at time $t$, every node of height at
least $k_0$ on their path is unpinned at time $t$, so
$\sum_iM_{v_i}(t)^2\le2(k_0+1)+2\sum_k(b\theta^2)^{k/2}=:\Phi_0$. By
Lemma~\ref{lem:pairmag}, $I(X_a;X_c\mid\mathcal F_t)\ge\tilde K\theta^{4j}$
with $\tilde K=\frac12(1-\theta^2)^2e^{-2K\Phi_0}$. Since $\E
N_{\mathcal H}\ge\frac34n$, Lemma~\ref{lem:collision} gives
\[
 \KL(P\,\|\,Q_A)\ge\tilde K\theta^{4j}\bigl(\tfrac34n-Dn/b^j\bigr).
\]
If $D\le b^j/2$, the divergence is at least $n\tilde K\theta^{4j}/4$.
Put $A=(n\tilde K/(4\eps))^{\alpha}$, so that $n\tilde K\theta^{4j}/4>\eps$
exactly when $b^j<A$. If $A\le1$ we use $D_\eps\ge1$. If $1<A\le n$, the
largest integer $j$ with $b^j<A$ satisfies $b^j\ge A/b$, so
$D_\eps>A/(2b)$. If $A>n$, we take $j=h$ and obtain $D_\eps>n/2$.
\end{proof}

\section{Gaussian Tree}\label{app:gauss}

This appendix proves Theorem~\ref{thm:gauss} and the Gaussian part of
Corollary~\ref{cor:peak}. Appendix~\ref{app:gauss-tools} gives the cost
and the pair correlations in terms of scalar messages,
Appendix~\ref{app:gauss-lower} the lower bounds and the upper bound up to
the threshold, and Appendix~\ref{app:gauss-above} the multistage schedule
above the threshold. Put $\eta=1-\rho^2$, $A=b\rho^2$ and
$g(z)=\rho^2z/(1+\eta z)$, with $g(\infty)=\rho^2/\eta$.

\subsection{Cost, Pairs, and Messages}\label{app:gauss-tools}

For disjoint $R,S$, let $C_{S\mid R}$ be the correlation matrix of
$X_S$ given $X_R$. The total correlation of a Gaussian batch is
$\TC(X_S\mid X_R)=-\frac12\log\det C_{S\mid R}$, and it does not depend on
the values $x_R$. A deterministic value-adaptive rule therefore incurs, on
each history, the cost of the ordered partition it realizes, and a fixed
schedule attains the minimum divergence among rules with at most $D$
rounds.

\begin{lemma}[Pairwise control]\label{lem:gauss-pair}
Suppose $X=M+\xi$, where $M$ is Gaussian and the $\xi_i$ are independent
Gaussian variables, independent of $M$, with $\Var X_i=1$ and
$\Var\xi_i\ge\eta>0$. Write $c_{ij}=\corr(X_i,X_j\mid X_R)$ and
$\Lambda=\max\{1,\lambda_{\max}(C_{S\mid R})\}$. Then
\[
 \frac1{2\Lambda}\sum_{i<j\in S}c_{ij}^2\ \le\ \TC(X_S\mid X_R)\ \le\
 \frac1{2\eta}\sum_{i<j\in S}c_{ij}^2 .
\]
\end{lemma}

\begin{proof}
The conditional covariance is $\Cov(M_S\mid X_R)+\operatorname{diag}(\Var\xi_i)$,
and its diagonal is at most $1$, so $C_{S\mid R}\succeq\eta I$. The
function $f(t)=t-1-\log t$ satisfies
$(t-1)^2/(2\max\{t,1\})\le f(t)\le(t-1)^2/(2\min\{t,1\})$. Apply it to the
eigenvalues of $C_{S\mid R}$ and use $\operatorname{tr}C_{S\mid R}=|S|$ and
$\|C_{S\mid R}-I\|_F^2=2\sum_{i<j}c_{ij}^2$.
\end{proof}

For an oriented edge $u\to v$, let $\lambda_{u\to v}(R)$ be the likelihood
precision supplied at $u$ by the generated leaves on the $u$ side of the
edge. These precisions satisfy
$\lambda_{u\to v}=\sum_{w\sim u,w\ne v}g(\lambda_{w\to u})$, with precision
$+\infty$ at a generated leaf and $0$ at an ungenerated one. For two
ungenerated leaves,
\begin{equation}\label{eq:gauss-path}
 \corr(X_i,X_j\mid X_R)=\prod_{uv\in\operatorname{path}(i,j)}r_{uv}(R),\qquad
 r_{uv}(R)=\frac{\rho}{\sqrt{[1+\eta\lambda_{u\to v}(R)][1+\eta\lambda_{v\to u}(R)]}} .
\end{equation}
Completing the square across an edge gives $g$, precisions from
conditionally independent branches add, and inverting the two-variable
precision matrix of an edge gives $r_{uv}$. Since $g$ is increasing, every
edge factor decreases as leaves are generated.

\subsection{Lower Bounds and the Upper Bound up to the Threshold}\label{app:gauss-lower}

\begin{lemma}[A uniform pair lower bound]\label{lem:gauss-floor}
Let $i,j\notin R$ have nearest common ancestor of height $d\ge1$. Then
\[
 \corr(X_i,X_j\mid X_R)\ge\frac{\rho^{2d}}{1+A}\Bigl(\frac{1-A}{1-A^d}\Bigr)^2,
\]
with the ratio read as $1/d$ when $A=1$. Consequently the correlation is at
least $K\rho^{2d}$ if $A<1$, at least $b^{-d}/(2d^2)$ if $A=1$, and at
least $K(b\rho)^{-2d}$ if $A>1$, with $K=(A-1)^2/(1+A)$.
\end{lemma}

\begin{proof}
Let $L_k$ be the precision at the root of a subtree of height $k$ when all
its leaves are generated. Then $L_0=\infty$, $L_{k+1}=bg(L_k)$, and
$L_k=(A-1)/(\eta(1-A^{-k}))$ for $A\ne1$, $L_k=1/(\eta k)$ for $A=1$. The
sequence decreases in $k$. Let $M_k$ be the precision at the parent of an
edge from height $k$ to height $k+1$ from all leaves outside the child's
subtree; then $M_k=(b-1)g(L_k)+g(M_{k+1})$ with $M_h=0$, and backward
induction gives $M_k\le L_{k+1}\le L_k$. The messages produced by any $R$
are at most these full-observation messages. Hence, for an edge between
heights $k$ and $k+1$ with $k\ge1$,
$r_e(R)\ge\rho/(1+\eta L_k)=\rho(1-A^k)/(1-A^{k+1})$, and the edge at an
ungenerated leaf has factor at least $\rho/\sqrt{1+A}$. Each half of the
path has one leaf edge and one edge at each height $1,\dots,d-1$; the
products telescope.
\end{proof}

The variance of the root explained by all leaves is $L_h/(1+L_h)$, whose
limit is positive exactly when $A>1$. This is the reconstruction threshold
of the Gaussian model.

\begin{proof}[Proof of the lower bounds in Theorem~\ref{thm:gauss}]
A Gaussian pair with correlation $r$ has mutual information
$-\frac12\log(1-r^2)\ge r^2/2$. Lemma~\ref{lem:gauss-floor} holds for every
conditioning set, so Lemma~\ref{lem:collision}, with all states admissible
and the subtrees of height $j$ as groups, gives
$\KL\ge\mu_j(n-Dn/b^j)$, where $\mu_j$ is the pair bound at height $j$.
If $n\mu_j\ge2\eps$, then $D_\eps\ge b^j/2$. For $A\ne1$ we have
$\mu_j=Bq^{4j}$ with $q=\rho$ when $A<1$ and $q=(b\rho)^{-1}$ when $A>1$.
Put $U=(nB/(2\eps))^{\log b/(4\log(1/q))}$ and
$j=\min\{h,\lfloor\log_bU\rfloor\}$ when $U\ge b$; this gives
$D_\eps\ge\min\{n,U\}/(2b)$. The exponent is $\alpha=1/(4x)$ for $A<1$ and
$\alpha_{\mathrm{lo}}=1/(4(1-x))$ for $A>1$, since $\log(b\rho)=(1-x)\log
b$. At $A=1$, use $\mu_j=b^{-2j}/(8j^4)$ and
$U=\sqrt{n/(16\eps)}/h^2$; the same choice of $j$ gives
$D_\eps\ge(8b)^{-1}\min\{n,\sqrt{n/\eps}/h^2\}$.
\end{proof}

\begin{lemma}[Uniform sparse schedule]\label{lem:gauss-uniform}
If $b\rho^4<1$, the schedule that generates one leaf of every subtree of
height $d$ per round has divergence at most $Cn\rho^{4d}$. Hence
$D_\eps\le C\min\{n,(n/\eps)^{1/(4x)}\}$.
\end{lemma}

\begin{proof}
For a leaf $i$ of a batch, at most $(b-1)b^{t-1}$ other batch leaves have
nearest common ancestor with $i$ at height $d+t$, and by~\eqref{eq:gauss-path}
their squared conditional correlations are at most $\rho^{4(d+t)}$ under
any earlier generation. The sum over the batch is at most
$(b-1)\rho^4\rho^{4d}/(1-b\rho^4)$. Lemma~\ref{lem:gauss-pair} with
innovation variance $\eta$ at every leaf bounds the total correlation of a
batch by a constant times $|S|\rho^{4d}$. Summing over the $n$ leaves and
choosing the smallest suitable $d$ proves the claim; if $d$ reaches $h$,
sequential generation is exact.
\end{proof}

This proves the upper bounds in (i) and (ii), because $b\rho^4\le1/b<1$
for $b\rho^2\le1$.

\subsection{Saturation and the Multistage Schedule}\label{app:gauss-above}

Suppose $A>1$. Put $\kappa=1/(1-2x)$, $\lambda_\ast=(A-1)/\eta$ and
$\beta=(b\rho)^{-1}$. Label each leaf by its sequence of child indices in
$\{0,\dots,b-1\}^h$, and let $P_s$ be the leaves whose last $s$ indices are
zero. Thus $P_s$ contains one leaf of every subtree of height $s$, and
$P_h\subset\dots\subset P_0$.

\begin{lemma}[Saturation]\label{lem:saturation}
There is $C$ such that, for $1\le s\le h$, every $R\supseteq P_s$ and every
$i,j\notin R$ with nearest common ancestor of height $t$,
\[
 \corr^2(X_i,X_j\mid X_R)\le C\pi_s(t),\qquad
 \pi_s(t)=\rho^{4\min(t,\kappa s)}\beta^{4(t-\kappa s)_+}.
\]
\end{lemma}

\begin{proof}
By monotonicity we may take $R=P_s$. For $k\ge s$ the upward precision
$\lambda_k$ at a node of height $k$ satisfies $\lambda_s=\rho^{2s}/(1-\rho^{2s})$
and $\lambda_{k+1}=bg(\lambda_k)$, whose solution is
\[
 \frac1{\lambda_k}=\frac1{\lambda_\ast}+A^{-(k-s)}\Bigl(\rho^{-2s}-1-\frac1{\lambda_\ast}\Bigr).
\]
Hence $(\lambda_\ast-\lambda_k)_+\le\lambda_\ast^2A^{-(k-s)}\rho^{-2s}=
\lambda_\ast^2A^{-(k-\kappa s)}$, using $A^s\rho^{-2s}=b^s=A^{\kappa s}$.
The parent-side precisions satisfy $m_k=(b-1)g(\lambda_k)+g(m_{k+1})$,
$m_h=0$. Since $\lambda_\ast=bg(\lambda_\ast)$ and $0\le g'\le\rho^2$,
$(\lambda_\ast-m_k)_+\le(b-1)\rho^2(\lambda_\ast-\lambda_k)_++\rho^2(\lambda_\ast-m_{k+1})_+$,
and expanding toward the root gives
$(\lambda_\ast-m_k)_+\le CA^{-(k-\kappa s)}+\lambda_\ast\rho^{2(h-k)}$.
Because $1+\eta\lambda_\ast=A$ and $\log(1+\eta z)$ has derivative at most
$\eta$, every edge at height $k\ge\kappa s$ satisfies
$r_e\le\beta\exp\{C(A^{-(k-\kappa s)}+\rho^{2(h-k)})\}$; below $\kappa s$
we use $r_e\le\rho$. Each half of the path has one edge at every height
$0,\dots,t-1$, and the error terms are summable uniformly in $s,t,h$.
\end{proof}

\emph{The schedule.} First generate the single leaf of $P_h$. Then, for
$s=h,h-1,\dots,1$, generate $N_s=P_{s-1}\setminus P_s$, which has
$(b-1)nb^{-s}$ leaves. At the start of stage $s$ all of $P_s$ is
generated. Choose $H_s\in[s,h]$; every subtree of height $H_s$ contains
$(b-1)b^{H_s-s}$ leaves of $N_s$, and we generate one of them per subtree
in each of $(b-1)b^{H_s-s}$ rounds. The total number of rounds is
$D=1+(b-1)\sum_sb^{H_s-s}$. A leaf of such a batch has at most
$b^{t-H_s}$ companions with nearest common ancestor at height $t>H_s$, so
by Lemmas~\ref{lem:saturation} and~\ref{lem:gauss-pair} stage $s$ costs at
most
\begin{equation}\label{eq:stage-cost}
 C|N_s|\sum_{t=H_s+1}^hb^{t-H_s}\pi_s(t).
\end{equation}

\begin{lemma}[Exponent calculation]\label{lem:exponent}
Let $0<x<1/2$, $y>1/4$, $\kappa=1/(1-2x)$ and
$a\ge\max\{2x/(1+2x),1/(4y)\}$. For $s\ge0$ and $M>0$ put $K=\kappa s$,
$H=s+aM$ and $\phi_s(t)=4x\min(t,K)+4y(t-K)_+$. Then
$\sup_{t\ge H}\{t-H-\phi_s(t)\}\le-(M-s)$.
\end{lemma}

\begin{proof}
The function $t-\phi_s(t)$ has slope $1-4x$ below $K$ and $1-4y<0$ above
$K$, so the supremum is attained at $t=H$ if $H\ge K$ or $x\ge1/4$, and at
$t=K$ otherwise. Note that $H\le K$ is equivalent to $s\ge
aM(1-2x)/(2x)$.

\emph{Supremum at $t=H\le K$.} We need $4x(s+aM)\ge M-s$, that is
$(1+4x)s\ge(1-4xa)M$. Using $s\ge aM(1-2x)/(2x)$, it suffices that
$a[(1-2x)(1+4x)+8x^2]/(2x)\ge1$, which is $a(1+2x)\ge2x$.

\emph{Supremum at $t=H>K$.} We need $4xK+4y(H-K)\ge M-s$, that is
$c_s\,s+(4ya-1)M\ge0$ with $c_s=1+4y+4\kappa(x-y)=(1+2x-8xy)/(1-2x)$. The
second term is nonnegative. If $c_s<0$, use $s<aM(1-2x)/(2x)$: it
suffices that $a(1+2x-8xy)/(2x)+4ya-1=a(1+2x)/(2x)-1\ge0$.

\emph{Supremum at $t=K>H$ with $x<1/4$.} We need $K-H-4xK\le-(M-s)$, which
after substituting $\kappa=1/(1-2x)$ reads $\kappa s\ge(1-a)M$. Since $s>
aM(1-2x)/(2x)$, it suffices that $a/(2x)\ge1-a$, that is $a(1+2x)\ge2x$.
\end{proof}

For the Gaussian model, $\pi_s(t)=b^{-\phi_s(t)}$ with $y=1-x$, so that
$1/(4y)=1/(4(1-x))$.

\begin{proof}[Proof of the upper bound in Theorem~\ref{thm:gauss}(iii)]
Let $M=\log_b(C_0nh^2/\eps)$ for a large constant $C_0$ and
$H_s=\min\{h,\lceil s+\alpha_{\mathrm{up}}M\rceil\}$. Since
$\pi_s(t)=b^{-\phi_s(t)}$ and $H\mapsto-H+\sup_{t\ge H}(t-\phi_s(t))$ is
nonincreasing, Lemma~\ref{lem:exponent} gives
$\sum_{t>H_s}b^{t-H_s}\pi_s(t)\le hb^{-(M-s)}$. By~\eqref{eq:stage-cost}
each stage costs at most $C(b-1)nhb^{-M}\le\eps/h$, and stages with
$H_s=h$ cost nothing, so the total divergence is at most $\eps$. Since
$H_s-s\le\alpha_{\mathrm{up}}M+1$, the number of rounds is at most
$1+(b-1)hb^{\alpha_{\mathrm{up}}M+1}\le Ch(nh^2/\eps)^{\alpha_{\mathrm{up}}}$.
\end{proof}

\begin{proof}[Proof of Corollary~\ref{cor:peak}, Gaussian part]
For fixed $\eps$ the factors involving $h$ contribute $o(1)$ to the
exponent of $n$. Below the threshold the exponent is $1/(4x)<1/2$, above it
both $\alpha_{\mathrm{lo}}$ and $\alpha_{\mathrm{up}}$ are smaller than
$1/2$, and at the threshold the exponent is $1/2$. All three expressions
tend to $1/2$ as $x\to1/2$.
\end{proof}

\section{Binary Tree Above the Threshold}\label{app:binary-above}

This appendix proves Theorem~\ref{thm:binary-above}.
Appendix~\ref{app:path} bounds the expected squared correlation of two
leaves along their path by a Bayesian backward induction,
Appendix~\ref{app:assembly} combines it with the schedule of
Appendix~\ref{app:gauss-above}, and Appendix~\ref{app:interval} describes
the computer-assisted verification of the one-step condition. The message
map is $f(z)=\operatorname{artanh}(\theta\tanh z)$, with $f'(z)=
\theta(1-t^2)/(1-\theta^2t^2)$ for $t=\tanh z$, and
$F=f'^2=\theta^2G(\tanh z)$ with $G$ as in Section~\ref{sec:above}.

\subsection{The Path Bound}\label{app:path}

Let $i,j$ be ungenerated leaves and $v_1,\dots,v_\ell$ the interior nodes of
their path, listed from $i$ to $j$. The \emph{side evidence} $E_k$ of
$v_k$ consists of the generated leaves in the components of the tree minus
$v_k$ that contain neither $v_{k-1}$ nor $v_{k+1}$. Let $h_k$ be the field
that $E_k$ exerts on $Z_{v_k}$, and define $u_1=0$, $\zeta_k=u_k+h_k$ and
$u_{k+1}=f(\zeta_k)$.

\begin{lemma}[Derivative representation]\label{lem:derivative}
With $m_i,m_j$ the conditional means of $X_i,X_j$,
\[
 \corr(X_i,X_j\mid x_R)=\theta\sqrt{\frac{1-m_j^2}{1-m_i^2}}\ \prod_{k=1}^\ell f'(\zeta_k),
\]
and hence
\[
 \corr^2(X_i,X_j\mid x_R)\le\frac{\theta^2}{1-\theta^2}\prod_{k=1}^\ell F(\zeta_k).
\]
\end{lemma}

\begin{proof}
Add an external field $\delta$ at the leaf $i$. By linear response,
$\Cov(X_i,X_j\mid x_R)=\partial_\delta\E[X_j\mid x_R]|_{\delta=0}$. Belief
propagation is exact on the tree, and the field travels only along the path,
so the chain rule gives $(1-m_j^2)f'(0)\prod_kf'(\zeta_k)$ with
$f'(0)=\theta$. Divide by $\sqrt{(1-m_i^2)(1-m_j^2)}$ and use $|m_i|\le\theta$.
\end{proof}

\begin{lemma}[Sequential structure]\label{lem:sequential}
Let $\mathcal G_k$ be generated by $X_{E_1},\dots,X_{E_k}$. Then $u_{k+1}$
is the half log-likelihood ratio of $\mathcal G_k$ for $Z_{v_{k+1}}$, and
$X_{E_{k+1}}$ is conditionally independent of $\mathcal G_k$ given
$Z_{v_{k+1}}$.
\end{lemma}

Both statements follow from the Markov property, because $E_1,\dots,E_k$
lie on one side of $v_{k+1}$. Let $\tau_k=\tanh h_k$, the posterior mean of
$Z_{v_k}$ given $X_{E_k}$ under a uniform prior, and let $\pi_k$ be its
law, which is symmetric. Put $\sigma_k=\E\tau_k^2$. Consistency of binary
symmetric experiments gives $\Pr(\tau\in d\tau\mid Z=\pm1)=(1\pm\tau)\pi(d\tau)$;
under a prior with mean $t$, $\tau$ has law $(1+t\tau)\pi(d\tau)$ and the
posterior mean is $(t+\tau)/(1+t\tau)$.

\begin{lemma}[One step]\label{lem:onestep}
For every symmetric $\pi$ with $\int\tau^2d\pi\ge y$ and every $|t|\le\theta$,
\[
 \theta^2\int(1+t\tau)\,Q_a\Bigl(\frac{t+\tau}{1+t\tau}\Bigr)\pi(d\tau)\le\lambda_a(y)\,w_a(t).
\]
\end{lemma}

\begin{proof}
By symmetry of $\pi$ the left side equals $\theta^2\int\Psi_a(t,|\tau|)\,
\pi(d\tau)$, the integral of a function of $\tau^2$. This function is at
most $\widehat\Psi_{a,t}(\tau^2)$, and $\widehat\Psi_{a,t}$ is concave, so
by Jensen's inequality the integral is at most
$\widehat\Psi_{a,t}(\int\tau^2d\pi)\le\sup_{y\le\tau^2\le1}\widehat\Psi_{a,t}(\tau^2)$.
The definition of $\lambda_a$ gives the claim.
\end{proof}

\begin{proposition}[Path bound]\label{prop:pathbound}
For every $a$ with $a\theta^2<1$,
\[
 \E\,\corr^2(X_i,X_j\mid X_R)\le\frac{\theta^2}{(1-\theta^2)(1-a\theta^2)}\prod_{k=1}^\ell\lambda_a(\sigma_k).
\]
\end{proposition}

\begin{proof}
By Lemma~\ref{lem:derivative} it suffices to bound $\E\prod_kF(\zeta_k)$.
Let $V_{\ell+1}\equiv1$ and
\[
 V_k(u)=\E_u\bigl[F(u+h_k)\,V_{k+1}(f(u+h_k))\bigr],
\]
where $h_k$ has its law under the prior field $u$. By
Lemma~\ref{lem:sequential},
$\E[\prod_{q\ge k}F(\zeta_q)\mid\mathcal G_{k-1}]=V_k(u_k)$. We show by
backward induction that
\[
 V_k(u)\le\frac{w_a(\tanh u)}{1-a\theta^2}\prod_{q\ge k}\lambda_a(\sigma_q).
\]
For $k=\ell+1$ this holds because $w_a\ge1-a\theta^2$. For the step, put
$t=\tanh u$ and $m=\tanh(u+h)=(t+\tau)/(1+t\tau)$. Since $\tanh
f(z)=\theta\tanh z$,
\[
 F(u+h)\,w_a\bigl(\tanh f(u+h)\bigr)=\theta^2G(m)(1-a\theta^2m^2)=\theta^2Q_a(m),
\]
and Lemma~\ref{lem:onestep} gives $\E_u[F\cdot w_a(\tanh
f(\cdot))]\le\lambda_a(\sigma_k)w_a(t)$. Finally $u_1=0$ and $w_a(0)=1$.
\end{proof}

The factors $F(\zeta_k)$ along a path are dependent, and the bound does not
factorize their expectation. Each step of the induction takes the
conditional expectation over the new side evidence only, and the evidence
behind enters through the prior field $u$, over which
Lemma~\ref{lem:onestep} takes the supremum. The weight $w_a$ credits the
information carried by the incoming message.

\begin{proposition}[Batch cost]\label{prop:batchpair}
Let $S$ be a batch, $i\in S$ with parent $p$, and $S\setminus i=\{j_1,\dots,j_r\}$
in any order. Then
\[
 \E\,I(X_i;X_{S\setminus i}\mid X_R)\le\frac{\theta^2}{1-\theta^2}\sum_{q=1}^r
 \E\,\corr^2\bigl(Z_p,X_{j_q}\mid X_{R\cup\{j_1,\dots,j_{q-1}\}}\bigr).
\]
\end{proposition}

\begin{proof}
We have $I\le I_{\chi^2}=\Var(\E[X_i\mid X_{S\setminus i},x_R]\mid
x_R)/\Var(X_i\mid x_R)$. Since $i$ is an ungenerated leaf with parent $p$,
$\E[X_i\mid\cdot]=\theta\E[Z_p\mid\cdot]$, and $\Var(X_i\mid x_R)\ge
1-\theta^2$. The Doob martingale $M_q=\E[Z_p\mid x_R,x_{j_1},\dots,x_{j_q}]$
has orthogonal increments. Because $X_{j_q}$ is binary,
$\E[Z_p\mid X_{j_q},\mathcal F_{q-1}]$ is affine in $X_{j_q}$, so the $q$th
increment has conditional second moment
$\corr^2(Z_p,X_{j_q}\mid\mathcal F_{q-1})\Var(Z_p\mid\mathcal F_{q-1})$.
\end{proof}

The correlation $\corr(Z_p,X_j\mid\cdot)$ has the representation of
Lemma~\ref{lem:derivative} from the side of $j$, with the interior nodes of
the path from $i$ to $j$ other than $p$, and Proposition~\ref{prop:pathbound}
applies to it.

\begin{lemma}[Linear comparison]\label{lem:linear}
For a node $v$ and a set $E$ of leaves, $\E\,\E[Z_v\mid X_E]^2$ in the
binary model is at least the squared multiple correlation of $Z_v$ with
$X_E$ in the Gaussian model with $\rho=\theta$.
\end{lemma}

\begin{proof}
Both models have unit variances and covariances
$\theta^{\operatorname{dist}(u,w)}$. The linear least-squares error depends
only on second moments and equals the Bayes error in the Gaussian model.
In the binary model the Bayes error $1-\E\,\E[Z_v\mid X_E]^2$ is at most the
linear error.
\end{proof}

\subsection{Assembly}\label{app:assembly}

Let $T=b\theta^2>1$. In the Gaussian model with $\rho=\theta$, a node whose
$b-1$ side subtrees are fully generated receives precision
$(b-1)g(\lambda_\ast)=\lambda_s=(b-1)(T-1)/(b-T)$, that is squared multiple
correlation $y^\ast_b=\lambda_s/(1+\lambda_s)$. In the proof of
Lemma~\ref{lem:saturation}, the upward precisions under $P_s$ converge to
$\lambda_\ast$ geometrically above height $\kappa s$. Hence, for every
$\epsilon'>0$ there is $c(\epsilon')$ such that every node of a path that
is not the nearest common ancestor and has height at least
$\kappa s+c(\epsilon')$ has Gaussian side correlation at least
$(1-\epsilon')y^\ast_b$ whenever $R\supseteq P_s$. By
Lemma~\ref{lem:linear}, $\sigma_k\ge(1-\epsilon')y^\ast_b$ for these nodes.

\begin{proof}[Proof of Theorem~\ref{thm:binary-above}]
The condition $b\lambda_b(\theta)<1$ is verified in
Appendix~\ref{app:interval}, and it gives $a_{\mathrm{bin}}(\theta)<1/2$.
Let $\gamma>a_{\mathrm{bin}}(\theta)$. The bound increases with $\gamma$,
so we may assume $\gamma<1/2$. Then $\lambda_b(\theta)<b^{-1/(2\gamma)}$,
and by the definition of $\lambda_b(\theta)$ there is an admissible $a$,
that is, $a\in[0,2)$ with $a\theta^2<1$ and $\lambda_a(0)\le\theta^2$, such
that $\lambda_a(y^\ast_b)<b^{-1/(2\gamma)}$. Since $0\le\Psi_a\le1$, each
$\widehat\Psi_{a,t}$ is concave with values in $[0,1]$, hence Lipschitz on
every compact subinterval of $(0,1)$ with a constant that does not depend
on $t$. As $w_a\ge1-a\theta^2>0$, the function $\lambda_a$ is continuous
on $(0,1)$, and we may choose $\epsilon'>0$ with
$\lambda:=\lambda_a((1-\epsilon')y^\ast_b)\le b^{-1/(2\gamma)}$. Put
$y=\log(1/\lambda)/(2\log b)$, so that $1/(4y)\le\gamma$ and $y>1/2$.

Use the schedule of Appendix~\ref{app:gauss-above}. In stage $s$ the
context contains $P_s$. For two batch leaves with nearest common ancestor
at height $t$, the path has two interior nodes at each height
$1,\dots,t-1$ and the ancestor itself. Nodes below height
$\kappa s+c(\epsilon')$, and the ancestor, contribute at most
$\lambda_a(0)\le\theta^2=b^{-2x}$ each; nodes above contribute at most
$\lambda=b^{-2y}$ each. By Propositions~\ref{prop:pathbound}
and~\ref{prop:batchpair}, the expected cost of a batch leaf is at most
$C\sum_{t}b^{t-H_s}b^{-\phi_s(t)}$ with $\phi_s(t)=4x\min(t,K)+4y(t-K)_+$
and $K=\kappa s+c(\epsilon')$, where the conditioning sets in
Proposition~\ref{prop:batchpair} contain $P_s$. The constant shift of $K$
changes the bound by a constant factor. Since
$\gamma\ge\max\{2x/(1+2x),1/(4y)\}$, Lemma~\ref{lem:exponent} applies with
$\gamma$ in place of $a$. With the choice of $M$ and $H_s$ in the proof of
Theorem~\ref{thm:gauss}(iii), and $\gamma$ in place of
$\alpha_{\mathrm{up}}$, the total divergence is at most $\eps$ and the
number of rounds is at most $C_\gamma h(nh^2/\eps)^{\gamma}$, where
$C_\gamma$ depends on $\gamma$ through $\epsilon'$ and $c(\epsilon')$.
\end{proof}

\subsection{Computer-Assisted Verification of the One-Step Condition}\label{app:interval}

It remains to show that some admissible $a$ satisfies (V0)
$\lambda_a(0)\le\theta^2$ and (V1) $b\lambda_a(y^\ast_b)<1$, for every
$b\ge2$ and $\theta\in[0.02,0.98]$ with $b\theta^2>1$.

\paragraph{Reduction.}
Put $R_a=\Psi_a/(1-\tau^2)$, a rational function,
\[
 R_a=\tfrac12(1-t^2)^2(1-\tau^2)\sum_\pm\frac{(1\pm t\tau)^2-a\theta^2(t\pm\tau)^2}{(1\pm t\tau)\,D_\pm^2},\qquad
 D_\pm=(1\pm t\tau)^2-\theta^2(t\pm\tau)^2,
\]
with $D_\pm>0$ on the domain. Both $\Psi_a$ and $R_a$ are even in $t$ and
in $\tau$.
\begin{itemize}
\item For $\theta\in[0.02,0.77]$ and all $b$, we verify
$R_a(t,\tau)\le w_a(t)$. Then $\Psi_a\le w_a$, which is (V0), and
$\lambda_a(y)\le\theta^2(1-y)$. Condition (V1) becomes
$g_b=T(b-T)/(T(b-2)+1)<1$, which is equivalent to $(T-1)^2>0$.
\item For $\theta\in[0.77,0.98]$ we verify (V0) as $\Psi_a\le w_a$. For
$b\in\{2,3,4\}$ we split the $t$-range into pieces and verify, on each, a
linear bound $\Psi_a\le w_a(t)(\alpha_j-\beta_j\tau^2)$ with $\beta_j\ge0$;
then $b\lambda_a(y^\ast_b)\le\max_jT(\alpha_j-\beta_jy^\ast_b)$. For $b\ge5$
we verify $R_a\le\alpha_jw_a$, so that $b\lambda_a(y^\ast_b)\le\max_j\alpha_j\,g_b\le\max_j\alpha_jg_5$,
because $g_b=b^2\theta^2(1-\theta^2)/(b(b-2)\theta^2+1)$ decreases in $b$ when
$b\theta^2>1$.
\item The inequalities are tight at the corner $(t,\tau)=(0,0)$, where the
gradient vanishes. There we use Taylor's theorem with Lagrange remainder
and verify that the interval Hessian is positive definite on a corner box.
This requires $2-1/\theta^2<a<2$; away from the corner all inequalities are
strict.
\end{itemize}

\paragraph{Implementation and result.}
The verification uses interval arithmetic in which every IEEE-754 binary64
operation is followed by outward rounding by one unit in the last place,
and a branch-and-bound search over boxes in $(\theta,t,\tau)$ that uses the
larger of the natural and the centered-form enclosures. The functions,
gradients and Hessians are generated symbolically. The parameter $a$ is a
multiple of $1/256$, and all constants $\alpha_j,\beta_j$ are stored binary64
numbers. The interval endpoints are binary64 numbers as well. The two
outermost intervals are $[0.0199,0.02]$ and $[0.9799,0.9801]$, whose
binary64 endpoints enclose the real numbers $0.02$ and $0.98$, so the
certificates cover the closed real interval $[0.02,0.98]$.
Table~\ref{tab:interval} summarizes the certificates.

\begin{table}[t]
\centering\small
\begin{tabular}{>{\raggedright\arraybackslash}p{0.2\textwidth}lrr>{\raggedright\arraybackslash}p{0.33\textwidth}}
\toprule
part & range of $\theta$ & intervals & boxes & result\\
\midrule
$R_a\le w_a$ & $[0.02,0.77]$ & 76 & $4.7\times10^4$ & all verified; smallest lower bound $5.2\times10^{-8}$\\
(V0) and linear bounds & $[0.77,0.98]$ & 43 & $5.2\times10^6$ & all verified; smallest (V0) margin $1.3\times10^{-9}$\\
(V1), recomputed & $[0.77,0.98]$ & 43 & -- & $\max b\lambda_a$ is $0.981$ ($b=2$), $0.796$ ($b=3$), $0.768$ ($b=4$), $0.959$ ($b\ge5$)\\
\bottomrule
\end{tabular}
\caption{Interval certificates for the one-step condition. The (V1) bounds
are recomputed in interval arithmetic from the stored certificates, with
$y^\ast_b$ evaluated at the lower end of each $\theta$-interval.}
\label{tab:interval}
\end{table}

For $\theta\in(0.98,0.999]$ a grid evaluation gives $b\lambda_a\le0.68$
for $b\in\{2,\dots,6,8\}$, but the number of boxes grows quickly as
$\theta\to1$ and we have not completed the interval verification there.

\section{Generating the Plan}\label{app:plan}

This appendix proves Theorems~\ref{thm:plan} and~\ref{prop:granularity}.

\begin{lemma}[Unseparated pairs]\label{lem:treedepth}
If all nodes of a complete $b$-ary subtree of height $H$ are generated in
at most $H$ rounds, then some round generates two nodes of the subtree
whose connecting path contains no node generated in an earlier round.
\end{lemma}

\begin{proof}
Induction on $H$. For $H=0$ there is nothing to prove, since a single node
cannot be generated in zero rounds. Consider the first round that generates
a node of the subtree. If it generates two nodes of the subtree, they form
the required pair, because nothing in the subtree was generated earlier and
the path between two nodes of a subtree stays in the subtree. Otherwise it
generates one node $v$. Some child of the root has a subtree of height
$H-1$ that does not contain $v$; all of its nodes are generated in strictly
later rounds, of which there are at most $H-1$. The induction hypothesis
gives the pair.
\end{proof}

\begin{proof}[Proof of Theorem~\ref{thm:plan}]
\emph{Upper bound.} Generate the root in round $1$ and the nodes of depth
$\ell$ in round $\ell+1$. Given the nodes of depth $\ell-1$, the nodes of
depth $\ell$ are conditionally independent, so every round has zero total
correlation and Proposition~\ref{prop:cost} gives an exact sample in $h+1$
rounds. This holds in both models.

\emph{Lower bound.} Let $a,c$ be an unseparated pair at distance $\ell$
generated in one round. Conditionally on everything generated before, the
path between them is a binary Markov chain of unrevealed nodes. Each
receiving node on the path has an external field of at most $bJ$ from its
neighbors off the path, because every message is at most $J$ in absolute
value; hence each edge slope is at least
$\beta_\ast=s(bJ)=\frac12[\tanh((b+1)J)-\tanh((b-1)J)]$. The endpoint $a$
has field at most $(b+1)J$, so $\Var X_a\ge1-\tanh^2((b+1)J)$. Pinsker's
inequality gives $I(X_a;X_c\mid x_R)\ge c_0\beta_\ast^{2\ell}$. Take the
$n/b^H$ disjoint subtrees whose roots have height $H$, and let $D\le H$.
On every run of the rule, Lemma~\ref{lem:treedepth} gives an unseparated
pair in each of them, at distance at most $2H$. In each round, given the
history, select in every subtree the first unseparated pair of the batch
in a fixed order, if there is one. The selected pairs are disjoint, so by
Lemma~\ref{lem:tc}(i) the total correlation of the round is at least
$c_0\beta_\ast^{4H}$ times their number. Every subtree has a selected pair
in some round of every run, and Proposition~\ref{prop:cost} gives a
divergence of at least $(n/b^H)c_0\beta_\ast^{4H}$. Taking $H=R$ gives the bound
$c_0n(\beta_\ast^4/b)^R$ for rules with at most $R\le h$ rounds.
Choosing the largest $H\le h$ for which this exceeds
$\eps$ gives $D^{\mathrm{all}}_\eps\ge H+1$, which is the bound.
\end{proof}

\begin{proof}[Proof of Theorem~\ref{prop:granularity}]
\emph{Upper bound.} Generate the nodes of height at least $s$ level by
level, in $h-s+1$ rounds with zero cost. Given the nodes of height $s$, the
$n/m$ subtrees below them are independent, and each is a broadcast tree of
height $s$ with a fixed root. Run the schedule of
Appendix~\ref{app:below-upper} with the same $d\le s$ in all of them
simultaneously. Lemma~\ref{lem:sparse} applies with a fixed root, so the
divergence is at most $Cn\theta^{4d}$, and the choice $b^d=\min\{m,
C'(n/\eps)^{\alpha}\}$ gives the bound; for $b^d=m$ the leaves are
generated one per round in each subtree and the cost is zero.

\emph{Lower bound, plan term.} The nodes of height at least $s$ form a
complete $b$-ary tree of height $H=h-s$ with $N=n/m=b^H$ nodes of height
$s$. Let a rule use at most $R\le H$ rounds, and take the $N/b^R$ disjoint
subtrees of this tree whose roots have height $s+R$. All their nodes are
generated, so Lemma~\ref{lem:treedepth} gives an unseparated pair in each
of them, at distance at most $2R$. The path between two plan nodes
consists of plan nodes, so no generated leaf lies on it. Generated leaves
off the path change the external fields, but every message is at most $J$
in absolute value, and the argument for Theorem~\ref{thm:plan} applies
verbatim. It gives
\[
 \KL(P\,\|\,Q_A)\ \ge\ c_0N(\beta_\ast^4/b)^R .
\]
Hence $D_\eps^{(s)}\ge\min\{H+1,\ \log(c_0b^H/\eps)/(\log b+4\log(1/\beta_\ast))\}$.
For $\eps\le1/2$ we have $\log(c_0b^H/\eps)\ge H\log b+\log(2c_0)$, which
is at least $\frac12H\log b$ when $H\ge H_0=2\log(1/(2c_0))/\log b$. For
$H<H_0$ we use $D_\eps^{(s)}\ge1$. In both cases $D_\eps^{(s)}\ge
c_1(H+1)=c_1(h-s+1)$ with $c_1>0$ depending only on $(b,\theta)$.

\emph{Lower bound, unit term.} We follow the proof in
Appendix~\ref{app:below-lower} and change only the bound on the
magnetizations. Let $v$ have height $k<s$, and let $u$ be its ancestor at
height $s$, at distance $s-k$. Its final magnetization is
$\E[Z_v\mid X,Z_{\mathrm{plan}}]$. Given $Z_u$, the observations outside
the subtree of $u$ are independent of $Z_v$, so the final magnetization is
$\E[Z_v\mid X_{L(u)},Z_u]$, where $L(u)$ is the set of leaves below $u$.
Lemma~\ref{lem:decay}, applied with $Z_u$ included, gives
$\E[\E[Z_v\mid X_{L(u)},Z_u]^2]\le C[(b\theta^2)^k+\theta^{2(s-k)}]$.
With the pinning threshold $\lambda_k^2=(b\theta^2)^{k/2}+\theta^{s-k}$,
Doob's inequality gives $\Pr(v\text{ ever pinned})\le
C[(b\theta^2)^{k/2}+\theta^{s-k}]$, since $(A^2+B^2)/(A+B)\le A+B$. Use
groups of height $j\le s-j_1$. The expected fraction of leaves with an
ever-pinned ancestor of height in $[k_0,j]$ is at most
$C(b\theta^2)^{k_0/2}+C\theta^{j_1}/(1-\theta)$, which is at most $1/4$ for
suitable $k_0,j_1$ depending only on $(b,\theta)$. Along the path of two
uncovered leaves, $\sum_iM_{v_i}^2\le C+C\theta^{j_1}/(1-\theta)$ is
bounded, so Lemma~\ref{lem:pairmag} again gives pairwise information at
least $\tilde K\theta^{4j}$. Plan nodes generated in the same round as
leaves only increase the total correlation of the round. As before,
$D\le b^j/2$ forces divergence at least $n\tilde K\theta^{4j}/4$, and the
choice $j=\min\{s-j_1,j_\eps\}$ gives
$D^{(s)}_\eps\ge\frac12\min\{b^{-j_1}m,b^{-1}(n\tilde K/(4\eps))^{\alpha}\}$.
When $s<j_1+k_0$, we have $m<b^{j_1+k_0}$, and $D^{(s)}_\eps\ge1$ gives
the same bound with a smaller constant. Hence $D^{(s)}_\eps\ge
c_2\min\{m,(n/\eps)^{\alpha}\}$.

\emph{Combination.} Since $\max\{u,w\}\ge(u+w)/2$, the two lower bounds
give $D^{(s)}_\eps\ge\frac12\min\{c_1,c_2\}\bigl[(h-s+1)+\min\{m,(n/\eps)^{\alpha}\}\bigr]$.
\end{proof}

\section{Experimental Details}\label{app:exp}

This appendix gives the settings of the experiments in
Section~\ref{sec:experiments}. All trees are binary ($b=2$), and the
leaves are indexed by their child-index sequences read as binary numbers,
so that $P_s$ of Appendix~\ref{app:gauss-above} is the set of positions
divisible by $2^s$.

\paragraph{Exact divergence.}
For a sample $x\sim P$ and a schedule, $\log P(x_L)$ is computed by the
upward recursion with the revealed nodes clamped, and the conditional
marginals of every round by the upward--downward recursion. A revealed
node $v$ is clamped by the factor $\exp\{B(x_vs-1)\}$ with $B=40$, which
changes probabilities by a factor $1+O(e^{-80})$. On trees with at most
$15$ nodes the log-probabilities and marginals agree with enumeration to
$4\cdot10^{-14}$. The divergence is the sample mean of $\log P(x_L)-\log
Q_A(x_L)$, and we report its standard error.

\paragraph{Depth.}
The divergence is computed for $D=1,2,4,\dots$ rounds until it falls below
$0.01$. The depth $D_\eps$ is obtained by linear interpolation of $\log
\KL$ against $\log D$. Between $D=n/2$ and $D=n$, where the divergence is
zero, a crossing is reported only as $D_\eps>n/2$. The exponent is the
least-squares slope of $\log_2D_\eps$ against $h$.

\paragraph{Schedules.}
With $D=2^k$ and $\mathrm{rev}_k$ the bit reversal of $k$-bit integers,
the stride schedule generates in round $j$ the positions congruent to
$\mathrm{rev}_k(j)$ modulo $D$. Each batch contains one leaf of every
subtree of height $k$, and the batches go from sparse to dense. The
random schedule draws a fresh uniform partition into $D$ batches of
sizes $\lfloor n/D\rfloor$ or $\lceil n/D\rceil$ for every sample. The
confidence schedule generates, in each round, the leaves with the largest
posterior $|\E[X_i\mid\text{revealed}]|$, with ties broken at random and
the same batch sizes.

The multistage schedules are those of Appendix~\ref{app:gauss-above}.
Stage $s$ generates $N_s=P_{s-1}\setminus P_s$ given exactly $P_s$ in
$2^{K_s}$ rounds, and round $r$ generates the positions
$q2^{s+K_s}+(2\,\mathrm{rev}_{K_s}(r)+1)2^{s-1}$, $q\ge0$, that is, one leaf
of $N_s$ in every subtree of height $s+K_s$. The stages $s>k$ may be
merged into one round that generates $P_k$. By the chain rule the
divergence is the sum of the cost of that round and the stage costs, and
we estimate each stage cost separately for every $K_s$, on two independent
sets of $512$ samples. The allocation $(k,K_1,\dots,K_k)$ minimizes
$\sum_s2^{K_s}+\lambda\cdot(\text{divergence})$ on the first set, over a
grid of $\lambda$, and the divergence is reported on the second set.
These computations use $h=6,\dots,13$. The stride schedule with $D=2^k$ is the member with $K_s=k-s$; on trees with
$h=3$ its divergence computed from the stage costs agrees with the direct
computation to $10^{-10}$.

\paragraph{Plans.}
The plan of granularity $g$ consists of the internal nodes of depth at
most $h-g$. They are generated level by level with exact conditionals,
which costs $h-g+1$ rounds and no divergence, and the leaves are then
generated by the stride schedule.

\paragraph{Gaussian channel.}
A round that generates $S$ given the revealed set $R$ costs
$\TC(X_S\mid X_R)=\frac12\bigl[\sum_{i\in S}\log\Var(X_i\mid
X_R)-\log\det\Cov(X_S\mid X_R)\bigr]$, which does not depend on the
values. By the chain rule, the terms $\log\det\Cov(X_S\mid X_R)$ of the
rounds of stage $s$ sum to $\log\det\Cov(X_{P_{s-1}})-\log\det\Cov(X_{P_s})$.
For a leaf set $A$ with complement $B$ in the tree,
$\log\det\Cov(X_A)=\log\det\Lambda_{BB}-\log\det\Lambda$, where $\Lambda$
is the precision matrix of all nodes. We compute $\log\det\Lambda_{BB}$ by
eliminating the nodes of $B$ from the leaves to the root, and the
conditional variances by a subsequent pass from the root to the leaves.
In every round of a multistage schedule the revealed leaves are periodic
with period $2^{s+K_s}$, and the nodes above height $s+K_s$ at equal depth
are equivalent. The computation therefore reduces to one subtree of
height $s+K_s$ and the path from its root to the root of the tree, at
cost $O(2^{s+K_s}+h)$ per round. For every stage, $K_s$ is increased until
the stage cost falls below $10^{-6}$, and larger $K_s$ are given cost
zero. This changes the divergence of any allocation by at most
$h\cdot10^{-6}$, and the limit of $2^{14}$ rounds per stage was never
reached. On trees with $h\le5$ the costs agree with computations from the
dense covariance matrix to $2\cdot10^{-14}$, and on trees with $h\le10$
the reduced computation agrees with the computation on the full tree to a
relative error of $2\cdot10^{-12}$. The allocation is chosen as in the
binary model; without sampling, one evaluation suffices. The depth
$D_\eps$ is interpolated as above, and the exponents are fitted over
$h=14,\dots,20$.

\paragraph{Comparison with the bounds.}
In the binary model above the threshold, Theorem~\ref{thm:binary-above}
bounds the depth by $C_\gamma h(nh^2/\eps)^{\gamma}$ for every
$\gamma>a_{\mathrm{bin}}$. For $\gamma$ close to $a_{\mathrm{bin}}$, the
factor $h^{1+2\gamma}$ adds between $0.22$ and $0.27$ to the local slope
of this bound over $h=8,\dots,13$, and all measured exponents of
Table~\ref{tab:exp-above} lie below this local slope. The selected
allocations give each stage between one and eight rounds, and in most
cases the subtree height grows with the stage, as in the proofs.

In the Gaussian model we evaluate local slopes at $h=17$, the middle of
the fitting range. At the threshold, the local slopes of the two bounds in
Theorem~\ref{thm:gauss}(ii) are $1/2-2/(17\log2)\approx0.33$ and $1/2$,
and the measured exponent $0.43$ lies between them. For $\rho\ge0.85$ we
have $\alpha_{\mathrm{lo}}=\alpha_{\mathrm{up}}$, and the factor
$h^{1+2\alpha_{\mathrm{up}}}$ in (iii) adds
$(1+2\alpha_{\mathrm{up}})/(17\log2)$, between $0.13$ and $0.14$, to the
local slope of the upper bound. The measured exponents exceed
$\alpha_{\mathrm{lo}}$ by $0.08$ to $0.09$, so they lie between the local
slopes of the two bounds.

\paragraph{Sample sizes and compute.}
The stride schedule and the plans use $1024$ samples per point, the
random and confidence schedules $512$, and each stage cost $2\times512$.
The exact computations in the binary model ran on $44$ to $56$ CPU
cores, and those in the Gaussian model took about three minutes on $64$
cores.

{\small\setlength{\bibsep}{0pt plus 0.5pt}
\bibliography{refs}}

\end{document}